\documentclass[sigconf]{acmart}
\AtBeginDocument{%
  }

\acmConference[ACM SIGSPATIAL'26]{34th ACM SIGSPATIAL International Conference on Advances in Geographic Information Systems}{Nov 03--06,
  2026}{Riverside, CA}

\copyrightyear{2026}
\acmYear{2026}
\setcopyright{cc}
\setcctype{by}
\acmConference[SIGSPATIAL '26]{The 34th ACM International Conference on Advances in Geographic Information Systems}{November 03--06, 2026}{Riverside, CA, USA}
\acmBooktitle{The 34th ACM International Conference on Advances in Geographic Information Systems (SIGSPATIAL '26), November 03--06, 2026, Riverside, CA, USA}
\acmDOI{10.1145/3841645.3844188}
\acmISBN{979-8-4007-2950-8/2026/11}

\begin{document}

\title{Seeing Less in a Sparse World: Cross-View Geolocalization from Limited Field-of-View Observations}


\author{Fahimul Aleem}
\affiliation{%
  \institution{University of Central Florida}
  \city{Orlando, FL}
  \country{United States}}
\email{fahimul.aleem@ucf.edu}

\author{Shruti Vyas}
\affiliation{%
  \institution{University of Central Florida}
  \city{Orlando, FL}
  \country{United States}}
\email{shruti@ucf.edu}

\renewcommand{\shortauthors}{Fahimul et al.}

\begin{abstract}
\textbf{Cross-view geolocalization} aims to determine the geographic location of a ground-level image by retrieving its corresponding satellite image, offering a promising alternative for localization in Global Navigation Satellite System (GNSS)-denied or degraded environments such as urban canyons, dense forests, and disaster scenarios. Despite recent progress, existing approaches remain limited by the \textbf{lack of globally diverse training datasets} and benchmarks that are often geographically restricted. Furthermore, most benchmarks rely on panoramic ground views, which do not reflect the \textbf{limited field-of-view imagery captured by real world cameras}.

To address these challenges, we introduce \textbf{Cross-View World 500k (CVW500k)}, a large-scale globally distributed dataset containing over 500,000 ground-satellite image pairs collected across diverse geographic regions and environments. Unlike previous datasets, CVW500k uses limited field-of-view ground images, providing a more realistic and challenging setting for cross-view geolocalization. In addition, we propose \textbf{GeoQueryNet}, a query-based cross-view fusion transformer framework designed to bridge the viewpoint gap between ground and satellite images. The model leverages a CLIP-based visual encoder with parameter-efficient LoRA adaptation and introduces a Cross-View Alignment Module that uses learnable query tokens to extract geographically consistent features across viewpoints.
Our code can be found \href{https://github.com/Fahim17/Cross_View_World_Dataset.git}{\color{blue} here}.
\end{abstract}




\begin{teaserfigure}
    \centering
    \includegraphics[width=0.99\textwidth]{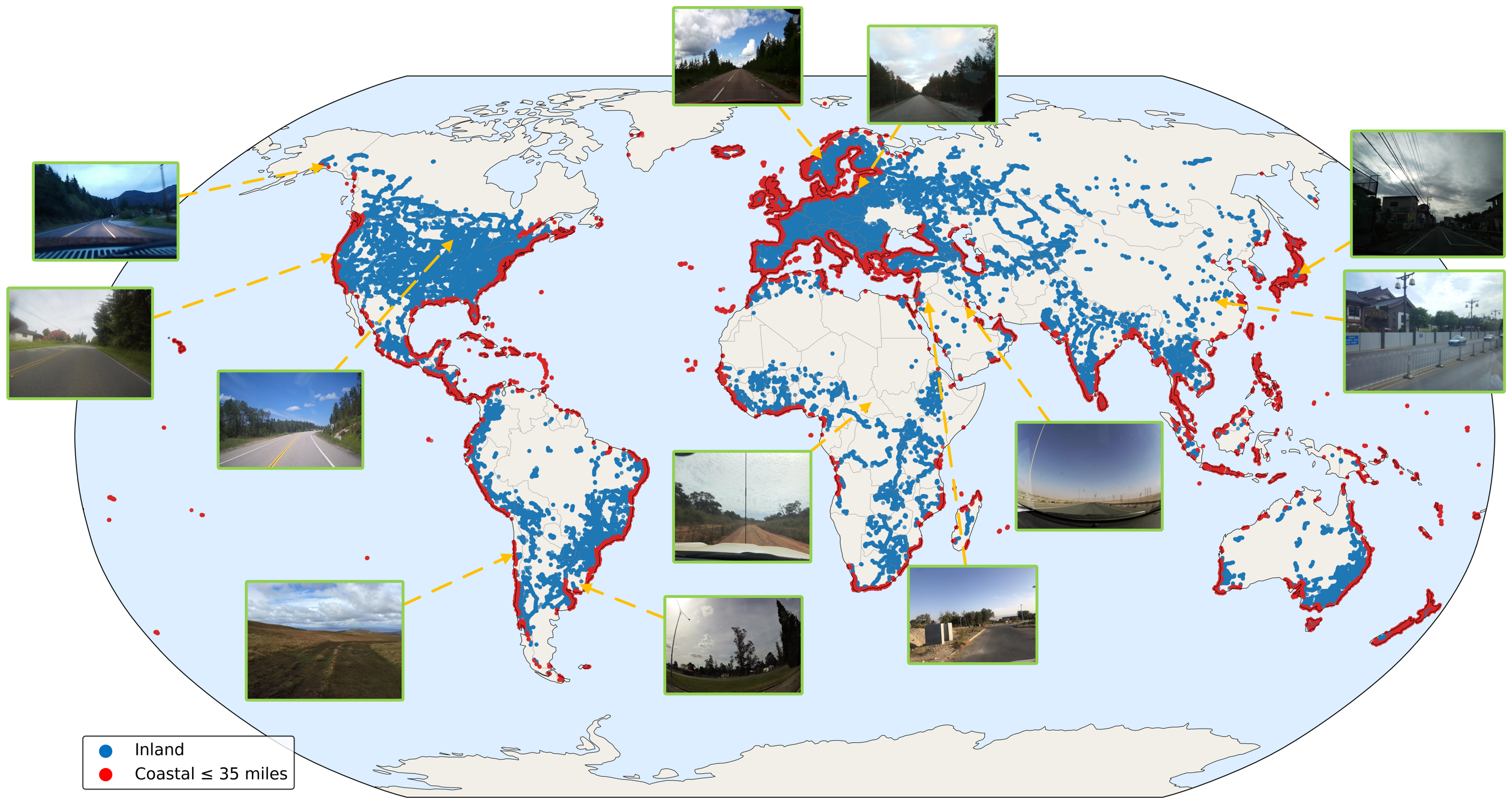}
    \caption{Global spatial distribution of CVW500k datapoints. \textcolor{blue}{BLUE} markers indicate inland locations, while \textcolor{red}{RED} markers represent coastal locations within 35 miles of the coastline. The overlaid ground-view examples illustrate the geographic and visual diversity of the dataset across different regions, road environments, landscapes, and urban-rural settings.}
    \label{fig:distribution}
\end{teaserfigure}

\maketitle

\section{Introduction}
\label{sec:intro}
Global Navigation Satellite System (GNSS) signals are highly susceptible to various forms of disruption such as trees, building, and electromagnetic interference\cite{huang2024cv}, which can significantly degrade positioning accuracy or lead to complete signal loss. These disruptions may arise from natural factors such as atmospheric disturbances, multipath reflections in urban canyons, or signal blockage in indoor, underground, or densely forested environments. In critical applications such as autonomous navigation, defense, and disaster response, such vulnerabilities pose serious operational risks. Vision-based localization, particularly cross-view geolocalization using ground and satellite imagery, has emerged as a promising complementary approach, capable of providing accurate and robust localization even in GNSS-denied or degraded environments. 

Cross-view geolocalization (CVGL) is the task of determining the geographic location of a ground-level image by matching it with a corresponding satellite or aerial image. This task is essential to a number of real-world applications, such as autonomous navigation\cite{xia2021cross} \cite{wang2022satellite} \cite{shi2022beyond}, augmented and virtual reality\cite{mithun2023cross}, and geo-spatial mapping and disaster mapping, emergency response\cite{zheng2020university}. By bridging the appearance gap between ground-level and overhead imagery, cross-view geolocalization enables machines to interpret spatial relationships and positioning beyond GPS constraints.

However, this problem remains highly challenging due to several factors. The first major difficulty is the extreme viewpoint and scale differences between ground-view and satellite-view images\cite{durgam2024cross}. The same location can appear drastically different when captured from a horizontal versus a top-down perspective-buildings, roads, and vegetation exhibit distinct geometric distortions, occlusions, and lighting conditions. Secondly, in real world scenario ground-view images typically a limited field of view, capturing only a portion of the surrounding environment. This narrow perspective limits the availability of global spatial cues visible in satellite imagery, making robust cross-view matching extremely difficult.

Recent advancements in vision foundation models have significantly improved the ability to learn generalized visual representations from large-scale and diverse image datasets. However, training visual geolocalization models based on such foundational architectures in global scale remains challenging due to the limited availability of globally distributed, high-quality cross-view datasets. Existing datasets such as CVUSA \cite{workman2015localize} and CVACT \cite{shi2020optimal}, VIGOR\cite{zhu2021vigor} which primarily focus on data from specific geographic regions. Such datasets do not adequately reflect the diversity of landscapes, architectures, and layouts found across the globe. The lack of large-scale, globally distributed datasets restricts the ability of models to generalize beyond local patterns. Moreover, many of these datasets are impractical for real-world applications, as their ground-view images are panoramic, which do not accurately represent the limited field-of-view (FoV) imagery typically captured by mobile devices or vehicle-mounted cameras. Consequently, models trained on one geographic region or dataset often fail to generalize to unseen cities or countries, as their learned representations are biased toward the visual characteristics of the training region.

To overcome the limitations of existing regional datasets and poor generalization in cross-view geolocalization, this paper introduces a globally distributed, realistic dataset and a novel learning framework for improved cross-view matching.
Our key contributions are as follows.

\begin{itemize}
    \item \textbf{Globally Realistic Dataset:} We introduce a large-scale, globally distributed sparse dataset, \textbf{Cross-View World 500k (CVW500k)}, that captures diverse geographic regions with limited field-of-view ground images, providing a more realistic and challenging cross-view geolocalization setting compared to existing regional or panoramic datasets. 
    

    \item \textbf{A Cross-View Framework:} We propose \textbf{GeoQueryNet} to align visual representations from different viewpoints, we propose a query-based cross-view alignment module that learns to extract geographically consistent information from two input images. The proposed approach improves the state-of-the-art performance on the world scale CVW500k dataset. 
    
\end{itemize}
\section{Related Work}
\label{sec:relatedwork}

\subsection{Cross-View Datasets}
\noindent
Over the past few years, several cross-view geolocalization datasets have been introduced to support research on aligning ground-level and aerial or satellite imagery. However, the majority of these datasets are region-specific, often covering only a limited number of urban areas. While such datasets have been instrumental in advancing model development and benchmarking, their restricted geographic scope limits generalization to unseen regions and diverse environments. 
\noindent
Among the most notable works, \textbf{CVUSA}\cite{workman2015localize}  is the most popular and widely used dataset, serves as a large-scale benchmark for cross-view geolocalization and place recognition research. It contains approximately 44 thousand ground-level images paired with corresponding satellite views, collected from 20 major U.S. cities, including New York, San Francisco, and Los Angeles. Each image pair is annotated with precise GPS coordinates and camera pose information, enabling accurate spatial correspondence. \textbf{CVACT}\cite{shi2020optimal}, another large-scale benchmark dataset specifically developed for cross-view geolocalization. Covering approximately 300 square miles around Canberra, Australia, the dataset comprises 128,334 ground-satellite image pairs, divided into 35,532 for training and 92,802 for testing. The ground-view images, collected from Google Street View, are 360° panoramas, while the satellite counterparts, obtained from Google Maps, are high-resolution images (1200 × 1200 pixels) with a ground resolution of 0.12 meters per pixel. A notable feature of CVACT is its inclusion of images captured under different lighting, weather, and seasonal conditions. Over time, CVUSA, together with CVACT, has become a standard benchmark for cross-view retrieval, primarily emphasizing one-to-one image matching. Its geographic and visual diversity makes it an essential resource for assessing the generalization ability of geolocalization models under real-world conditions.
\noindent
The \textbf{VIGOR}\cite{zhu2021vigor} dataset was designed to support fine-grained localization and capture real-world complexities that are not fully represented in traditional retrieval-based benchmarks. It contains aerial images and street-level panoramas collected from four major U.S. cities-New York, San Francisco, Chicago, and Seattle-with aerial images at 640 $\times$ 640 pixels and ground panoramas at 2048 $\times$ 1024 pixels. Unlike CVUSA and CVACT, which mainly assume one-to-one image retrieval between paired ground and aerial views, VIGOR introduces a more practical many-to-many formulation, where a single ground-view image may correspond to multiple nearby aerial images. The dataset follows a consistent spatial sampling interval of roughly 30 meters between panorama locations, enabling more precise neighborhood-level evaluation. It also deliberately includes 4\% unmatched aerial images as distractors, increasing retrieval difficulty and reducing reliance on simple pair matching. Additionally, each ground panorama is orientation-aligned with North at the center, providing standardized spatial referencing for fine-grained geolocalization evaluation.

\subsection{Cross-view Geolocalization}
CVGL has attracted significant attention in recent years due to its broad applicability in visual navigation, urban mapping, and autonomous driving. Over time, the research community has proposed a series of methods ranging from traditional handcrafted approaches to advanced deep learning frameworks that bridge this cross-view gap.

\noindent
\textbf{Deep Learning-Based Feature Embedding}
The emergence of deep learning marked a turning point in CVGL. Deep convolutional neural networks (CNNs) enabled end-to-end learning of discriminative and viewpoint-invariant features. Lin et al. \cite{lin2015learning} and Zhai et al. \cite{zhai2017predicting} demonstrated that CNNs trained with contrastive or triplet losses could significantly enhance the retrieval accuracy between ground and aerial views. Other approaches, \cite{shi2019spatial}, \cite{shi2020where}, \cite{yang2021cross}, and \cite{hu2022beyond}, utilized polar transformations to reshape satellite images, simulating a ground-level viewpoint and improving feature alignment. 

\noindent
\textbf{Transformer-Based and Contrastive Learning Frameworks}
Recent advances have shifted toward transformer-based architectures, which can capture long-range dependencies and modality correlations more effectively than CNNs. TransGeo\cite{zhu2022transgeo}, a transformer-based cross-view geolocalization model that leverages multi-head self-attention to align semantic structures between views. Transformers allow for flexible fusion of global context and local spatial features, significantly improving retrieval precision. Deuser et al. \cite{deuser2023sample4geo} proposed a contrastive learning framework with symmetric InfoNCE loss and a streamlined training pipeline, eliminating the need for aggregation modules and additional pre-processing. Pillai et al. \cite{pillai2024garet} introduced GAReT, a fully transformer-based framework for cross-view video geolocalization that eliminates the need for explicit viewpoint information such as odometry or camera intrinsics. Comparative analysis across model families reveals that models like Qwen3-VL leads in zero-shot accuracy and geographic coherence despite an unusual inverted scaling pattern, whereas InternVL2.5 follows traditional monotonic scaling and LLaVA variants demonstrate significantly weaker geographic grounding\cite{bharadwaj2026vision}.

\subsection{Why Do We Need the CVW500k Dataset?}
Existing cross-view geolocalization datasets such as CVUSA, CVACT, and VIGOR have significantly advanced research in matching ground-level images with aerial or satellite views. However, these datasets are geographically limited and do not fully capture the diversity of real-world environments. For example, CVUSA and CVACT are primarily collected from specific regions-mainly the United States and Australia-which introduces geographic bias and restricts a model's ability to generalize to unseen global locations. Similarly, VIGOR focuses on a limited number of cities, emphasizing urban environments while providing relatively little representation of rural, suburban, and geographically diverse regions. Another important limitation of these datasets is the lack of cultural landscape diversity, such as variations in architecture, road infrastructure, signage styles, vegetation patterns, and settlement layouts that differ significantly across countries and regions. Another dataset, GAMa\cite{vyas2022gama} dataset is a large-scale collection specifically designed for the novel task of cross-view video geolocalization. The work provides a large-scale benchmark dataset and proposes GAMa-Net with a hierarchical strategy for clip-level and video-level geolocalization, but the dataset is geographically concentrated in a few cities, has incomplete GPS labels and unaligned views, overlapping aerial images. However, the \textbf{CVW500k }dataset, which contains 500,000 satellite-ground image pairs collected worldwide, covers diverse continents, countries, climates, and cultural landscapes. By incorporating this broader geographic and cultural diversity, the dataset enables more realistic training and evaluation of geolocalization models and encourages the development of approaches that generalize effectively across different environments and regional characteristics worldwide.

\section{CVW500k dataset}
\label{sec:dataset}


\paragraph{Data Collection.}
CVW500k  is specifically designed to enhance the diversity and global coverage of cross-view geolocalization research. The ground-view images are sourced from the OpenStreetView5M dataset \cite{osv5m}, a publicly available dataset comprise of over 5.1 million geo-referenced street-view images spanning 225 countries and territories. Since OpenStreetView5M provides only ground-view images and does not include paired satellite reference images, OpenStreetView5M is intended for VLM based geolocalization evaluation rather than retrieval based cross-view matching. The dataset images are sourced from crowd-sourced platforms such as Mapillary\cite{mapillary}, which provides a globally diverse collection of open-access street-view imagery encompassing a wide range of environments-from densely populated urban areas to suburban regions and remote landscapes. These images have been widely utilized in constructing benchmark datasets for various computer vision tasks beyond geolocalization, semantic segmentation\cite{neuhold2017mapillary}, traffic sign detection and classification\cite{ertler2020mapillary}, place recognition\cite{warburg2020mapillary}, and visual localization\cite{jafarzadeh2021crowddriven}. From this dataset, we randomly selected about 500,000 street-view images and retrieved their corresponding satellite-view counterparts using the Google Maps API at zoom level 20, ensuring spatial alignment and diversity across various global environments. We divided the dataset into 90/10 split, where 438,606 ground-satellite image pairs in the training split and 48,735 pairs in the test split. Furthermore, to facilitate region-specific evaluation, we include an additional regional test subset containing 12,659 image pairs from Australia and New Zealand.
Table \ref{tab:dataset_comp} compares major Cross-View GeoLocalization (CVGL) datasets, highlighting how CVW500k advances beyond existing benchmarks. While CVUSA, CVACT, and VIGOR are limited to city-scale coverage with panoramic ground images, CVW500k provides global coverage with over 500,000 ground-satellite pairs, making it the largest and most diverse dataset. Unlike previous datasets that rely on panoramic views, CVW500k uses limited field-of-view (FoV) ground images, better reflecting real-world conditions.
Figure\ref{fig:distribution} illustrates the global geographic distribution of image pairs in the CVW500k dataset, highlighting its extensive and diverse coverage across continents. As shown, the dataset encompasses widespread representation across North America, Europe, Asia, and Africa, with particularly dense concentrations of samples in the United States, Western Europe, South America, and East Asia. Regions in South Asia, Africa, and Oceania are also well represented, ensuring coverage across both urban and rural landscapes. This broad spatial distribution demonstrates the global scalability and diversity of CVW500k, distinguishing it from prior CVGL datasets that are typically restricted to a few cities or countries. The worldwide spread of samples enhances the dataset's value for training and evaluating geolocalization models capable of generalizing across varying geographic regions, environmental contexts, and visual domains.

\begin{table*}[t]
\centering
\caption{Comparison between the proposed CVW500k dataset and existing datasets for cross-view geolocalization. Details are provided in the supplementary material.}
\begin{tabular}{lrrrrr}
\hline
\textbf{}                 & \textbf{CVUSA}                                               & \textbf{CVACT}                                               & \textbf{VIGOR}                                               & \textbf{GAMa}                                                & \textbf{CVW500k}                                             \\ \hline
\textbf{Viewpoints}       & \begin{tabular}[c]{@{}c@{}}Ground, \\ Satellite\end{tabular} & \begin{tabular}[c]{@{}c@{}}Ground, \\ Satellite\end{tabular} & \begin{tabular}[c]{@{}c@{}}Ground, \\ Satellite\end{tabular} & \begin{tabular}[c]{@{}c@{}}Ground, \\ Satellite\end{tabular} & \begin{tabular}[c]{@{}c@{}}Ground, \\ Satellite\end{tabular} \\
\textbf{Area Coverage}    & City                                                         & City                                                         & City                                                         & City                                                         & World wide                                                   \\
\textbf{Satellite images} & 44,416                                                       & 128,334                                                      & 90,618                                                       & 1.97M                                                        & 500,098                                                      \\
\textbf{Ground images}    & 44,416                                                       & 128,334                                                      & 238,696                                                      & 51,535                                                       & 500,098                                                      \\
\textbf{Multiple Cities}  & Yes                                                          & No                                                           & Yes                                                          & Yes                                                          & Yes                                                          \\
\textbf{Field of View}    & Panorama                                                     & Panorama                                                     & Panorama                                                     & Limited                                                      & Limited                                                      \\
\textbf{Sat. resolution}  & 750$\times$750                                               & 1200$\times$1200                                             & 640$\times$640                                               & 256$\times$256                                               & 640$\times$640                                               \\
\textbf{Sampling}         & Discrete                                                     & Dense                                                        & Dense                                                        & Discrete                                                     & Discrete                                                     \\ \hline
\end{tabular}
\label{tab:dataset_comp}
\end{table*}
\section{Methodology}
\label{sec:method}

\paragraph{Problem statement.}
Cross-view geolocalization aims to estimate the location of a ground-view image by retrieving its corresponding satellite-view image from a large gallery. Formally, let $\mathcal{D} = \{(I_g^i, I_s^i)\}_{i=1}^{N}$ be a dataset containing $N$ pairs of ground-view images $I_g \in \mathbb{R}^{H_g \times W_g \times 3}$ and satellite-view images $I_s \in \mathbb{R}^{H_s \times W_s \times 3}$, where each $(I_g^i, I_s^i)$ forms a positive pair corresponding to the same geographic location. The goal is to learn an embedding function $f: \mathbb{R}^{H \times W \times 3} \to \mathbb{R}^d$ that maps images into a $d$-dimensional feature space, ensuring that for each ground-view query $I_g^i$, the embedding $f(I_g^i)$ is close to its corresponding satellite image embedding $f(I_s^i)$ while being distant from embeddings of non-matching satellite images.

\subsection{Overview of GeoQueryNet}

We propose \textbf{GeoQueryNet}, a query-based cross-view fusion transformer network for Ge\textbf{o}-localization. GeoQueryNet effectively align cross-view visual representations between satellite and ground-level images. The framework consists of three main components: a Satellite image encoder, a Ground image encoder, and a  Cross-view Alignment module.

\subsection{Vision Encoder}
We utilize the CLIP visual encoder~\cite{radford2021learning} to derive feature embeddings from both ground-view and satellite-view images. Using a shared encoder ensures that both modalities are mapped into a common feature space, which facilitates subsequent alignment of their representations. To preserve the generalization ability of CLIP image encoder while adapting it for geolocalization, we adopt LoRA~\cite{hu2022lora}, a parameter-efficient fine-tuning technique that introduces low-rank trainable matrices into the attention layers rather than updating the entire set of model parameters. With a rank of (r = 16) and a dropout rate of 0.05, this approach significantly reduces the number of trainable parameters, mitigates overfitting, and maintains the valuable pre-trained knowledge of CLIP. Both the satellite and ground image encoders extract high-level visual features from their respective viewpoints, capturing distinct scene structures and spatial cues. We define $V_g: \mathbb{R}^{H_g \times W_g \times 3} \to \mathbb{R}^{d}$ as the ground view encoder and $V_s: \mathbb{R}^{H_s \times W_s \times 3} \to \mathbb{R}^{d}$ as the satellite view encoder, where $d$ is the feature embedding dimension. For a given ground-view image ( $I_g$ ), its encoded representation is obtained as $v_g = V_g(I_g)$ , and for a satellite image $I_s$, the corresponding feature embedding is $v_s = V_s(I_s)$. These feature vectors are then utilized for feature alignment across the two modalities.

\subsection{Cross-View Alignment Module}

To bridge the large viewpoint gap between ground-level and satellite imagery, we introduce a \textbf{Cross-View Alignment Module} that learns to integrate information from the two views through query-driven attention. Unlike standard feature matching, this module is designed to handle \textbf{limited field-of-view ground images} that do not fully overlap with satellite images; only a small region in the satellite view corresponds to the ground image. The module employs a set of learnable query tokens that act as intermediate representations to extract geographically consistent features from both perspectives.

\paragraph{Cross-View Query Alignment}

To integrate information from both views, we introduce a set of learnable query tokens:

\begin{equation}
Q \in \mathbb{R}^{K \times d}
\end{equation}

where $K$ is the number of queries. These queries act as latent alignment variables that iteratively aggregate information from both feature embeddings through cross-attention.

First, the queries attend to the ground-view embedding:

\begin{equation}
Q^{(1)} = \text{CrossAttn}(Q, v_g)
\end{equation}

This allows the queries to capture the most informative cues visible in the limited ground perspective, such as local road layouts, buildings, and vegetation.

Next, the updated queries attend to the satellite-view embedding:

\begin{equation}
Q^{(2)} = \text{CrossAttn}(Q^{(1)}, v_s)
\end{equation}

Through this interaction, the queries integrate complementary information from the satellite image, focusing on the region corresponding to the ground view. Stacking multiple transformer layers progressively refines the cross-view representation.

\paragraph{Cross-View Representation}

The final query embeddings encode features that are consistent across the two viewpoints. 

The learnable queries act as geographic anchors, selectively extracting features from both views that are useful for localization. The cross-view alignment module encourages the network to focus on \textbf{structural patterns that are consistent across viewpoints}, even when the ground image has a limited field of view and only partially overlaps with the satellite image. By using query tokens as an intermediate representation, the model selectively extracts and reconciles information from both views, enabling robust cross-view matching under partial overlap and sparse geographic coverage conditions.

\subsection{Learning objective}
To effectively align the feature representations of ground and satellite images, we employ a contrastive loss function, specifically the InfoNCE loss\cite{oord2018representation}, which is widely used in cross-view and multi-modal representation learning. The objective of this loss is to maximize the similarity between corresponding (positive) ground-satellite pairs while minimizing it for non-matching (negative) pairs within a batch.
Given a batch of $B$ ground-view images and their corresponding satellite-view images, the standard contrastive loss (InfoNCE) is formulated as:
\begin{equation}
    \mathcal{L}_{\text{InfoNCE}} = - \sum_{i=1}^{B} \log \frac{\exp(\text{sim}(\tilde{v}_g^i, \tilde{v}_s^i)/\tau)}{\sum_{j=1}^{B} \exp(\text{sim}(\tilde{v}_g^i, \tilde{v}_s^j)/\tau)}
\end{equation}
Here, each ground-view embedding $\tilde{v}_g^i$ is trained to be closer to its corresponding satellite embedding $\tilde{v}_s^i$, while being pushed away from all other satellite images $\tilde{v}_g^j$.
The InfoNCE loss encourages the model to assign a higher similarity score to the true match compared to all other mismatched pairs.


\section{Experiments and results}
\subsection{Datasets}

\begin{figure*}[t]
    \centering
    \includegraphics[width=0.99\textwidth]{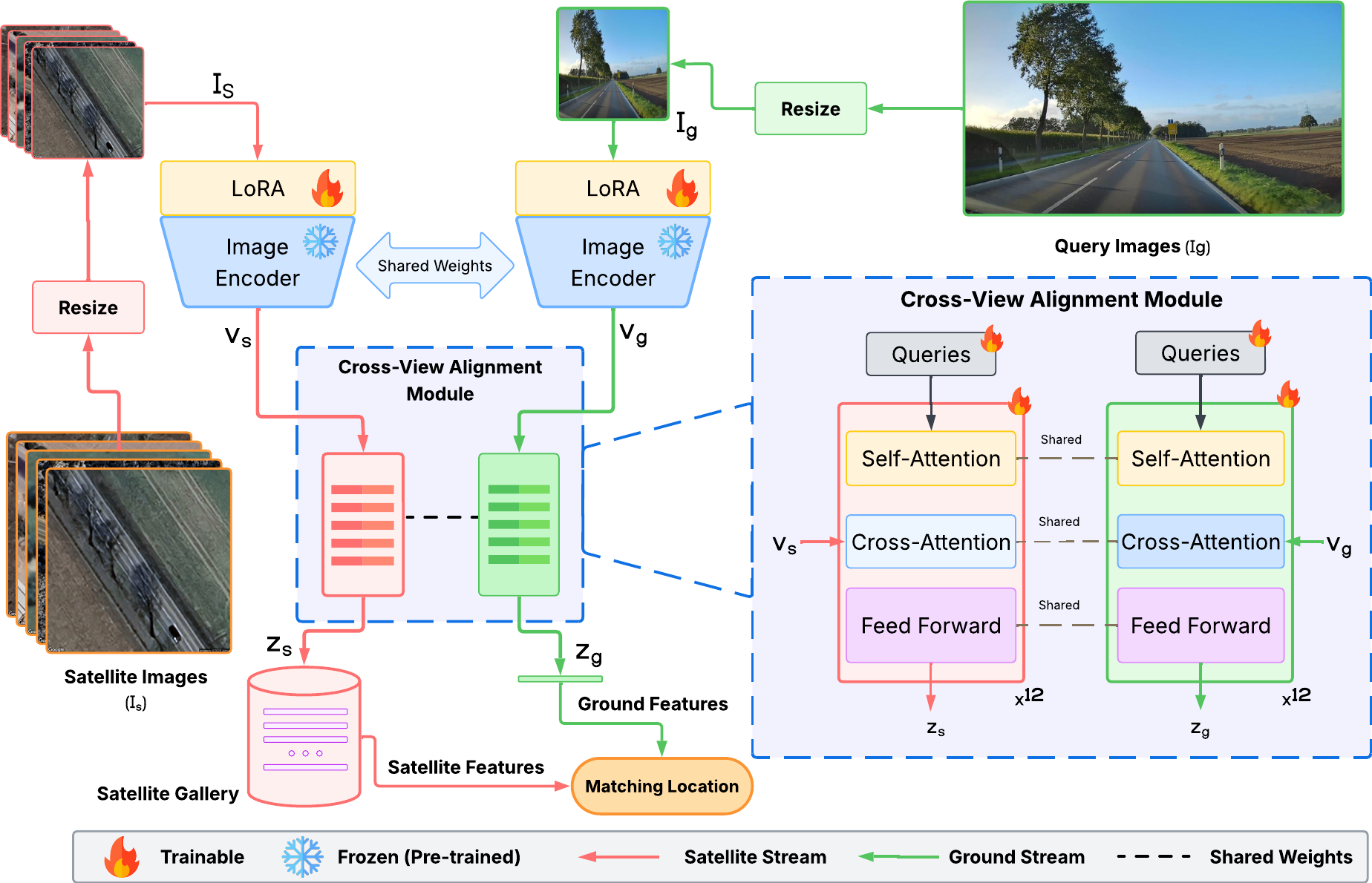}
    \caption{\textbf{GeoQueryNet}: Overview of the proposed GeoQueryNet framework. Ground and satellite images are processed through shared-weight image encoders with LoRA-based adaptation, producing view-specific features that are refined by the Cross-View Alignment Module. The aligned ground and satellite embeddings are then matched against the satellite gallery to retrieve the most likely geographic location.}
    \label{fig:placeholder}
\end{figure*}


We conduct our baseline experiments on three cross-view geolocalization datasets: \textbf{CVUSA}~\cite{workman2015wide}, \textbf{CVACT}~\cite{shi2020optimal}, and our curated dataset \textbf{CVW500k}. Each dataset consists of ground-satellite image pairs and is used to evaluate cross-view geolocalization performance under a unified experimental setting. The \textbf{CVUSA} dataset contains 44,416 ground-satellite image pairs, with 80\% used for training and the remaining 20\% used for testing. Similarly, the \textbf{CVACT} dataset has a comparable number of image pairs and also includes an additional extended set of 92,802 pairs; however, in our experiments, we use only the small version consisting of 44,415 ground-satellite image pairs. In addition to these standard benchmarks, we evaluate the baseline methods on our curated \textbf{CVW500k} dataset, which contains approximately 500K ground-satellite image pairs. Specifically, CVW500k is split into 438,607 training pairs and 48,736 testing pairs. These experiments allow us to compare the baseline methods across both widely used public benchmarks and our large-scale curated dataset.


\paragraph{Limited FoV:} As the original CVUSA and CVACT datasets provide panoramic ground-view images, we employ a random cropping strategy to simulate limited field-of-view (FoV) conditions. In this process, random horizontal segments are extracted from each panoramic image while preserving the original vertical resolution, thereby generating \textbf{90°} FoV ground-view samples along the azimuth direction. In contrast, the ground-view images in our CVW500k dataset are inherently limited-FoV, eliminating the need for additional cropping or preprocessing. More detail in supplementary.

\subsection{Evaluation metrics}
We adopt the standard Top-k Recall metric, a widely accepted evaluation measure in retrieval-based tasks such as cross-view geolocalization. This metric quantifies the proportion of query images for which the correct corresponding match is found within the top-k retrieved candidates. In our evaluation, we report Top-1, Top-5, Top-10, and Top-1\% Recall scores. The Top-1 metric indicating cases where the correct match is ranked first among all candidates. Top-5 and Top-10 provide a more flexible assessment by verifying whether the correct match appears within the top 5 or 10 retrievals. The Top-1\% metric is especially important for large-scale datasets, as it evaluates whether the correct match is retrieved within the top 1\% of the entire reference database.

\subsection{Training details}
We train our model using the q-former from Blip2 backbone. The embedded dimension is 768. The model is trained until performance converges, with the checkpoint achieving the highest validation accuracy retained for evaluation. We used Adam optimizer with a learning rate of $10^{-5}$. The model was fine-tuned using LoRA with a rank of 32 and a LoRA scaling factor of 32. A dropout rate of 0.1 was used for LoRA regularization, and no bias parameters were adapted. A single NVIDIA RTX A6000 GPU with a VRAM of 48GB was used. All of the experimental code has been written in Python.

\begin{table*}[t]
\centering
\caption{Comparison of baseline cross-view geolocalization methods on CVUSA, CVACT, and the proposed CVW500k dataset. All methods are fine-tuned under the same training protocol.}
\begin{tabular}{lrrrrrrrrrrrrrr}
\toprule
\multicolumn{1}{c}{\textbf{Methods}} & \multicolumn{4}{c}{\textbf{CVUSA}}                                                                                                           & \multicolumn{1}{c}{\textbf{}} & \multicolumn{4}{c}{\textbf{CVACT}}                                                                                                           & \multicolumn{1}{c}{\textbf{}} & \multicolumn{4}{c}{\textbf{CVW500k}}                                                                                                         \\ \cline{2-5} \cline{7-10} \cline{12-15} 
\multicolumn{1}{c}{\textbf{}}        & \multicolumn{1}{c}{\textbf{R@1}} & \multicolumn{1}{c}{\textbf{R@5}} & \multicolumn{1}{c}{\textbf{R@10}} & \multicolumn{1}{c}{\textbf{R@1\%}} & \multicolumn{1}{l}{}          & \multicolumn{1}{c}{\textbf{R@1}} & \multicolumn{1}{c}{\textbf{R@5}} & \multicolumn{1}{c}{\textbf{R@10}} & \multicolumn{1}{c}{\textbf{R@1\%}} & \multicolumn{1}{l}{}          & \multicolumn{1}{c}{\textbf{R@1}} & \multicolumn{1}{c}{\textbf{R@5}} & \multicolumn{1}{c}{\textbf{R@10}} & \multicolumn{1}{c}{\textbf{R@1\%}} \\ \hline
Sample4Geo\cite{deuser2023sample4geo}                           & 55.10                            & 78.30                            & 85.0                              & 96.6                               &                               & 27.90                            & 52.0                             & 62.30                             & 87.0                               &                               & 0.22                             & 0.86                             & 1.52                              & 23.98                              \\
GeoDTR\cite{zhang2023geodtr}                               & 13.20                            & 30.33                            & 39.58                             & 73.58                              &                               & 5.3                              & 15.67                            & 22.95                             & 57.64                              &                               & 2.01                             & 6.95                             & 11.08                             & 70.44                              \\
GeoDTR+\cite{zhang2024geodtrp}                              & 8.47                             & 22.60                            & 31.58                             & 69.48                              &                               & 6.04                             & 16.79                            & 24.94                             & 60.58                              &                               & 2.63                             & 8.71                             & 13.81                             & 75.45                              \\
ConGeo\cite{mi2024congeo}                               & 55.50                            & 75.40                            & 81.50                             & 93.90                              &                               & 40.60                            & 62.60                            & 69.80                             & 86.60                              &                               & 4.80                             & 13.35                            & 18.97                             & 69.87                              \\
MEAN\cite{mean}                                 & 5.23                             & 15.33                            & 22.74                             & 60.15                              &                               & 5.38                             & 15.87                            & 23.36                             & 61.09                              &                               & 3.45                             & 11.38                            & 17.43                             & 80.96                              \\
\bottomrule
\end{tabular}
\label{tab:models_on_datasets}
\end{table*}

\begin{table}[]
\centering
\caption{Performance comparison on the CVW500k dataset for GeoQueryNet and existing works. }
\begin{tabular}{lrrrr}
\toprule
\textbf{CVW500k}                                   & \textbf{R@1}             & \textbf{R@5}              & \textbf{R@10}             & \textbf{R@1\%}            \\ \midrule
Sample4Geo\cite{deuser2023sample4geo} & 0.22                     & 0.86                      & 1.52                      & 23.98                     \\
GeoDTR\cite{zhang2023geodtr}          & 2.01                     & 6.95                      & 11.08                     & 70.44                     \\
GeoDTR+\cite{zhang2024geodtrp} & 2.63 & 8.71  & 13.81 & 75.45 \\
ConGeo\cite{mi2024congeo}             & 4.80                     & 13.35                     & 18.97                     & 69.87                     \\
MEAN\cite{mean}                       & 3.45                     & 11.38                     & 17.43                     & 80.96                     \\
CAMP\cite{wu2024camp}                 & 5.47                     & 15.87                     & 23.38                     & 85.07                     \\
DSTG\cite{liang2025dstg}              & 0.01                     & 0.03                      & 0.05                      & 2.49                      \\
QDFL\cite{qdfl}                   & 8.19 &19.75 & 26.82 & 38.32 \\
SDPL\cite{chen2024sdpl}           & 2.30 & 6.77 & 9.98 & 47.02 \\
\textbf{GeoQueryNet}         & 15.19   & 42.10  & 54.04  & 90.87 \\ 
\bottomrule
\end{tabular}
\label{table:CVW500kresults}
\end{table}

\subsection{Baseline}

To evaluate the effectiveness of the proposed CVW500k dataset, we conducted a comprehensive baseline study using representative cross-view geolocalization methods. All baseline models were \textbf{fine-tuned} on CVW500k under a unified experimental protocol and trained until performance converges. The selected baselines cover various architectural designs and learning strategies: Sample4Geo\cite{deuser2023sample4geo} uses a weight-shared ConvNeXt-B Siamese encoder with symmetric InfoNCE loss and hard-negative sampling; GeoDTR\cite{zhang2023geodtr} uses a ResNet34 CNN backbone with a transformer-based geometric layout extractor to disentangle spatial layout from low-level visual features; GeoDTR+\cite{zhang2024geodtrp} extends this idea with an enhanced geometric layout extractor, contrastive hard sample generation, and a ConvNeXt-Tiny backbone; ConGeo\cite{mi2024congeo} introduces a model-agnostic contrastive objective for improving robustness to orientation and field-of-view variations, commonly using Sample4Geo/ConvNeXt-B as its base model; QDFL\cite{qdfl} adopts an adapted DINOv2-B vision transformer with query-driven feature learning through AQEU and FFU modules; 
CAMP\cite{wu2024camp} employs a ConvNeXt-B backbone together with contrastive attribute mining and position-aware partitioning to strengthen scene-level discrimination by combining global descriptors with fine-grained spatial cues. MEAN\cite{mean} adopts a lightweight ConvNeXt-Tiny backbone and introduces multi-level embedding, global-to-local feature enhancement, and cross-domain alignment branches to improve efficiency while preserving cross-view consistency. DSTG\cite{liang2025dstg} is built on Swin Transformer V2, where an STG teacher transfers knowledge to a distilled DSTG student through multiscale logit standardization for efficient matching. Collectively, these baselines cover CNN-based, transformer-based, contrastive, alignment-focused, and lightweight distillation designs, enabling a comprehensive evaluation of CVW500k against recent state-of-the-art cross-view geolocalization trends under a unified protocol.

\subsection{Results}


Table \ref{tab:models_on_datasets} shows that CVW500k introduces a substantially more challenging setting in the limited FoV evaluation than existing CVGL benchmarks. Although several methods achieve strong performance on CVUSA and CVACT, their Top-1 accuracy drops sharply on CVW500k, with the best-performing method, ConGeo, reaching only 4.80\% R@1. This indicates that CVW500k stresses the fine-grained retrieval capability of existing cross-view geolocalization models. At the same time, some models obtain high R@1\% on CVW500k, such as MEAN: 80.96\%, GeoDTR+: 75.45\%, and GeoDTR: 70.44\%. This means the correct match often appears somewhere in the broad candidate list, but not in the very top ranks. Therefore, these models are not completely failing to learn geographic similarity; rather, they fail at fine-grained place-level discrimination.

\noindent\textbf{Why these models fail on CVW500k?}
These models fail because most of them were designed around relatively controlled cross-view benchmarks, where the number of candidate locations is smaller and the visual domain is narrower. Sample4Geo uses a weight-shared ConvNeXt-B Siamese network with symmetric InfoNCE and hard-negative sampling, but its simple global embedding may not capture enough fine-grained geographic cues for a 500k-scale global dataset. GeoDTR and GeoDTR+ improve geometric layout modeling, and GeoDTR+ was motivated by the need to improve cross-area generalization and avoid overfitting to low-level details; however, geometric layout alone cannot resolve cases where many locations share similar road patterns, vegetation, buildings, or urban structures. MEAN's high R@1\% but low R@1 suggests that multi-level embedding and alignment can place the correct candidate in a broad neighborhood of the ranking, but its lightweight ConvNeXt-Tiny backbone and alignment branches are still insufficient for precise top-rank discrimination under large-scale ambiguity.

\noindent
Table~\ref{table:CVW500kresults} compares the retrieval performance of \textbf{GeoQueryNet} with existing methods on the CVW500k dataset. GeoQueryNet achieves the best performance across all evaluation metrics, demonstrating a substantial improvement over prior approaches in both top-rank retrieval accuracy and broader candidate retrieval quality.
For strict top-1 retrieval, GeoQueryNet obtains an R@1 of 14.92\%, which is significantly higher than all competing methods. The strongest baseline under this metric is QDFL, with an R@1 of 8.19\%. GeoQueryNet therefore improves R@1 by 6.73 percentage points, corresponding to an approximately 82.17\% relative improvement over the best existing result. This indicates that GeoQueryNet is considerably more effective at ranking the correct geolocation as the first retrieved match.
The improvement becomes even more pronounced at higher recall levels. GeoQueryNet achieves R@5 and R@10 scores of 36.51\% and 48.50\%, respectively. Compared with QDFL, the best baseline for both R@5 and R@10, GeoQueryNet improves performance by 16.76 percentage points in R@5 and 21.68 percentage points in R@10. These gains suggest that GeoQueryNet not only improves the top prediction but also produces a more reliable ranked candidate list, placing the correct match within the top retrieved results more consistently than previous methods.


Overall, GeoQueryNet shows substantially advanced cross-view geolocalization performance on CVW500k. Its large gains in R@1, R@5, and R@10 highlight its ability to retrieve more accurate matches at the top of the ranking, while its R@1\% performance confirms its robustness in large-scale retrieval settings.

\begin{figure}
    \centering
    \includegraphics[width=1\linewidth]{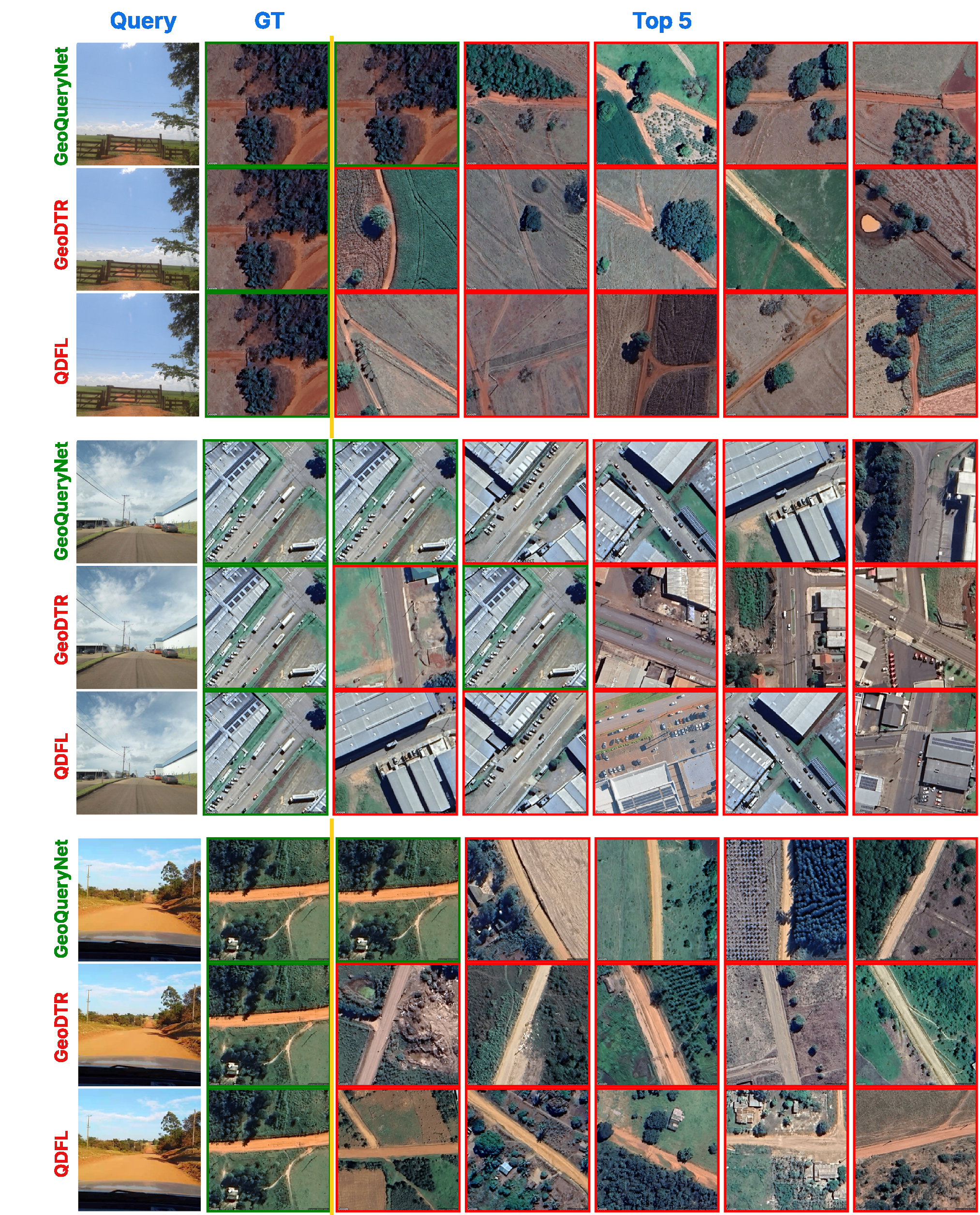}
    \caption{Qualitative comparison of cross-view geolocalization retrieval by GeoQueryNet, GeoDTR, and QDFL. Each group includes a ground-view query, ground-truth satellite image, and Top-5 retrieved satellite candidates per model.}
    \label{fig:qualitative_model}
\end{figure}

A key reason for the relatively poor performance of the other methods lies also in the data characteristics and training assumptions used during their development. Most existing approaches including GeoDTR, Sample4Geo, and DSTG were primarily trained and evaluated on datasets composed of panoramic ground images, where a 360° field of view provides significantly richer contextual information such as road layouts, building distributions, and surrounding landmarks. In contrast, the CVW500k dataset contains limited field view for ground images rather than panoramas, which reduces the available contextual cues and makes cross-view matching more challenging. As a result, models designed to exploit panoramic context struggle to generalize effectively when only limited directional visual information is available. Among the baselines, ConGeo is the only method not explicitly designed around panoramic imagery, which explains why it performs comparatively better than the other baseline methods.

Another major challenge comes from the \textbf{scale} and \textbf{geographic diversity }of the dataset. Unlike earlier benchmarks that focus on limited geographic regions or a small number of cities, CVW500k operates at a global scale, covering diverse countries, environments, and cultural landscapes. This dramatically increases \textbf{intra-class variability} and reduces the likelihood of visually similar satellite-ground correspondences, making retrieval significantly more difficult. Variations in architecture, road infrastructure, vegetation, climate, and urban layout across different regions further complicate the matching process. Consequently, methods trained on regionally constrained datasets fail to generalize effectively to this world-scale retrieval problem, highlighting the importance of models that can capture more robust and globally consistent cross-view representations.

Figure \ref{fig:qualitative_model} demonstrate the comparative advantage of GeoQueryNet in identifying the correct satellite match from ground-view queries. For each selected test sample, GeoQueryNet retrieves the ground-truth satellite image at Rank-1, shown by the green border, while GeoDTR and QDFL fail to retrieve the correct match at the top rank. Although the incorrect retrievals from GeoDTR and QDFL often contain visually similar structures, such as rural roads, vegetation patterns, open fields, or industrial layouts, they correspond to different geographic locations. This highlights the difficulty of cross-view geolocalization, where many satellite regions share similar visual cues. In contrast, GeoQueryNet better preserves query-specific spatial and contextual information, enabling more accurate matching between the ground-view image and its corresponding satellite-view counterpart.

\subsection{Ablations}

\begin{figure}
    \centering
    \includegraphics[width=0.99\linewidth]{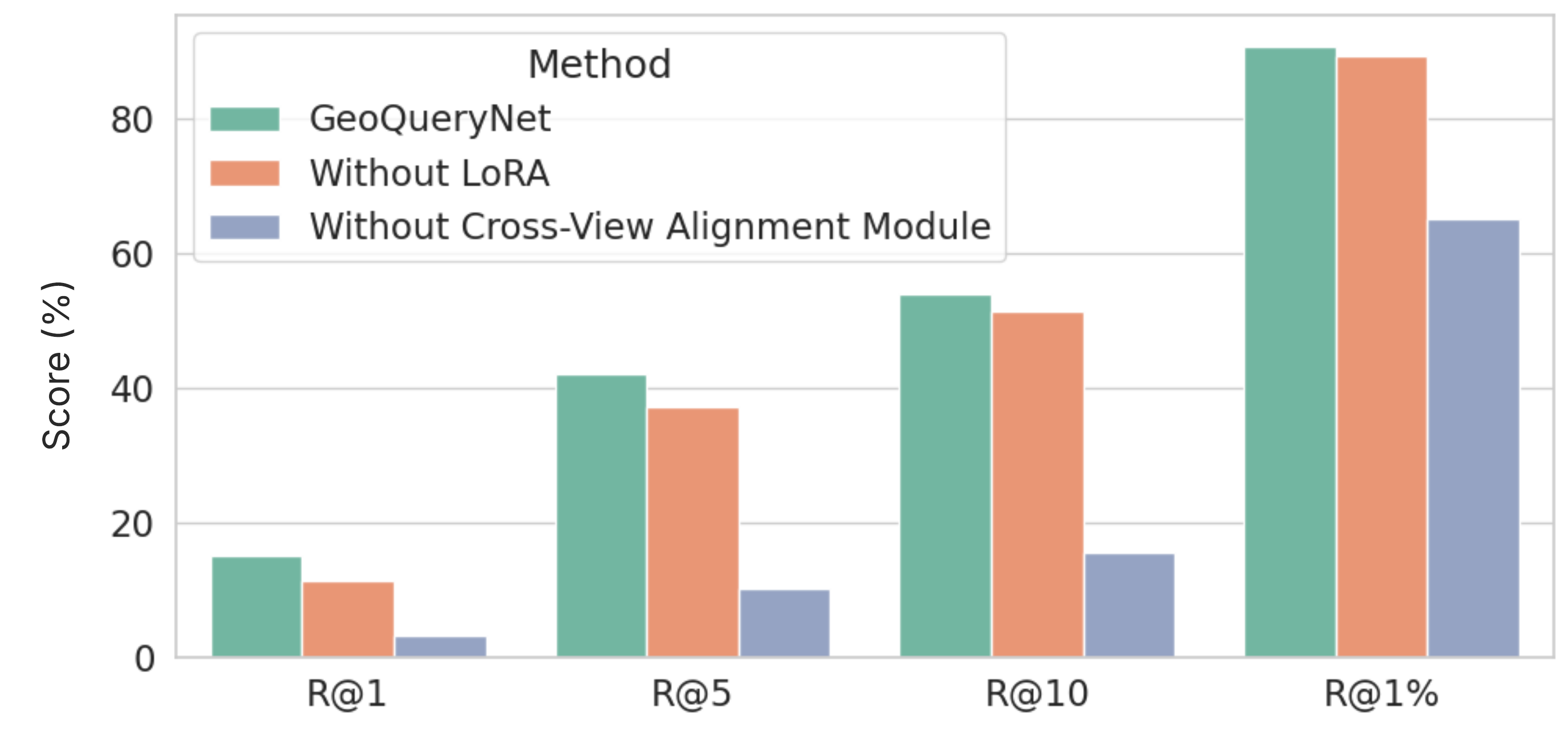}
    \caption{The impact of \textbf{LoRA adaptation} and the \textbf{Cross-View Alignment Module} on geolocalization retrieval performance. Results compare CVW500k retrieval scores when (Orange) LoRA is removed and only the alignment module is trained, and when (Blue) LoRA is used without the alignment module.}
    \label{fig:w_out_lora_cvam}
\end{figure}

Figure \ref{fig:w_out_lora_cvam} presents an ablation analysis to understand the individual contributions of LoRA-based adaptation in the image encoder and the Cross-View Alignment Module in the proposed framework. Two configurations are evaluated: (i) \textbf{without LoRA}, where only the Cross-View Alignment Module is trained, and (ii) \textbf{without the Cross-View Alignment Module}, where LoRA is applied to the encoder but cross-view alignment is removed. LoRA significantly reduces retrieval performance, even when the Cross-View Alignment Module is present. In this setting, the model achieves 11.36\% R@1, 37.11\% R@5, 51.33\% R@10, and 89.40\% R@1\%, indicating that the alignment module alone cannot fully adapt the pretrained encoder to the cross-view geolocalization task.

In contrast, when LoRA is used but the Cross-View Alignment Module is removed, performance drops more severely, with scores of 3.27\% R@1, 10.15\% R@5, 15.48\% R@10, and 65.04\% R@1\%. This result highlights that while LoRA helps adapt the encoder features, the absence of explicit cross-view alignment prevents effective matching between ground and satellite representations.

\begin{figure}
    \centering
    \includegraphics[width=1\linewidth]{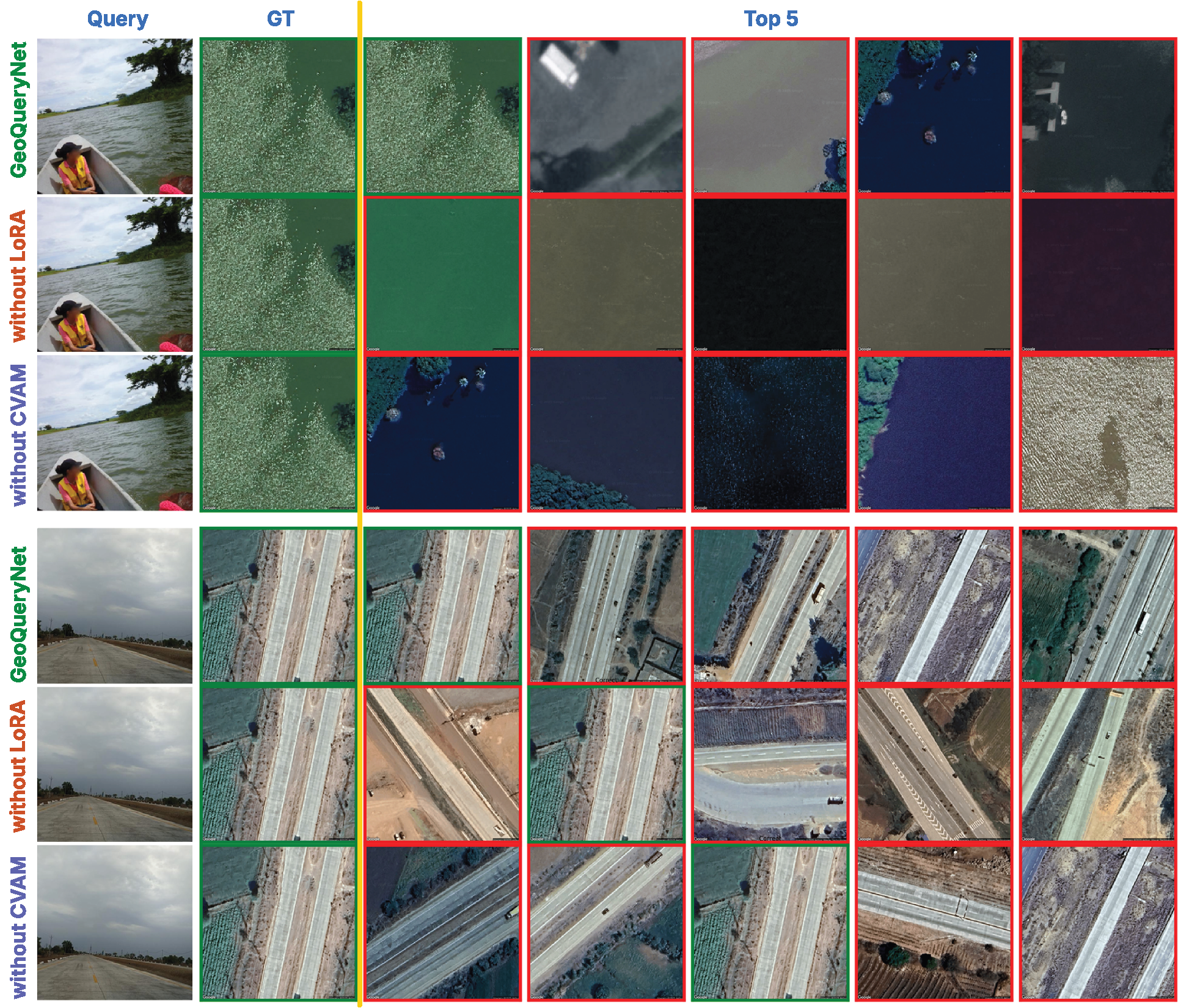}
    \caption{Qualitative ablation comparison of GeoQueryNet, GeoQueryNet without LoRA, and GeoQueryNet without Cross-View Alignment Module (CVAM).}
    \label{fig:GQN_LoRA_CVAM}
\end{figure}
Figure \ref{fig:GQN_LoRA_CVAM} show full GeoQueryNet consistently retrieves the ground-truth satellite image at Rank-1, showing strong alignment between ground-view and satellite-view features. Without LoRA, the model becomes less adapted to the cross-view geolocalization domain, while removing Cross-View Alignment Module (CVAM) weakens the embedding correspondence between the two views. As a result, the ablated variants retrieve visually similar but geographically incorrect satellite images.

Overall, the comparison demonstrates that both LoRA-based encoder adaptation and the Cross-View Alignment Module are critical components of the proposed framework, and their combination is necessary to achieve optimal cross-view geolocalization performance. Additional qualitative results are provided in the supplementary material.

\begin{table}[h]
\centering
\caption{LoRA rank ablation study showing the retrieval performance of GeoQueryNet on the CVW500k dataset using different LoRA ranks.}
\begin{tabular}{ccccc}
\hline
\textbf{Rank} & \textbf{Top 1} & \textbf{Top 5} & \textbf{Top 10} & \textbf{Top 1\%} \\ \hline
8             & 9.97           & 26.63          & 36.98           & 87.56            \\
16            & 10.45          & 27.60          & 37.99           & 89.00            \\
32            & 15.19   & 42.10  & 54.04  & 90.87          \\ 
64            & 13.41          & 33.78          & 45.49           & 96.06            \\ \hline
\end{tabular}
\label{tab:lora_rank_ablation}
\end{table}

\noindent\textbf{LoRA Rank Ablation.}
Table \ref{tab:lora_rank_ablation} presents an ablation study on the effect of LoRA rank on the retrieval performance of GeoQueryNet on the CVW500k dataset. The results indicate that increasing the LoRA rank from 8 to 16 yields only marginal improvements across all retrieval metrics, with Top-1 accuracy increasing from 9.97\% to 10.45\%. A more substantial performance gain is observed when the rank is increased to 32, achieving the best Top-1, Top-5, and Top-10 accuracies of 15.19\%, 42.10\%, and 54.04\%, respectively. This suggests that a rank of 32 provides sufficient representational capacity for adapting the model to the cross-view geolocalization task. However, further increasing the rank to 64 results in a decline in Top-1, Top-5, and Top-10 performance, indicating that additional parameters do not necessarily translate to better retrieval accuracy and may lead to overfitting or less effective optimization. Overall, the results demonstrate that a LoRA rank of 32 offers the best balance between model adaptability and retrieval accuracy, making it the optimal configuration for GeoQueryNet on the CVW500k dataset.

\begin{table}[]
\caption{Country-wise cross-view geolocalization retrieval performance measured using Top-5 and Top-10 accuracy. Train and Test column shows the Training and Test split for individual experiment.}
\centering
\begin{tabular}{lrrrrr}
\toprule
\textbf{GeoQueryNet} & \multicolumn{1}{l}{\textbf{T@1}} & \multicolumn{1}{c}{\textbf{T@5}} & \multicolumn{1}{c}{\textbf{T@10}} & \multicolumn{1}{c}{\textbf{Train}} & \textbf{Test} \\
\midrule
Argentina            & 26.09                            & 55.85                            & 68.81                             & 40561                              & 4521          \\
Colombia             & 28.91                            & 57.15                            & 67.47                             & 24860                              & 2890          \\
Chile                & 32.58                            & 59.04                            & 69.47                             & 22434                              & 2563          \\
Bolivia              & 36.85                            & 63.68                            & 74.64                             & 16958                              & 1881          \\
United States        & 36.91                            & 62.94                            & 74.74                             & 16745                              & 2178          \\
Russia               & 32.17                            & 61.59                            & 72.46                             & 12465                              & 1318          \\
Australia            & 49.21                            & 85.03                            & 92.71                             & 9132                               & 1016          \\
Canada               & 38.81                            & 67.11                            & 79.29                             & 7900                               & 868           \\
Ghana                & 34.76                            & 69.76                            & 81.66                             & 7609                               & 796           \\
India                & 30.23                            & 66.77                            & 81.40                             & 5400                               & 602           \\
China                & 39.10                            & 72.77                            & 81.68                             & 3679                               & 346           \\
Sierra Leone         & 45.38                            & 76.15                            & 86.15                             & 2337                               & 259           \\
Indonesia            & 40.83                            & 79.16                            & 91.66                             & 2148                               & 240           \\
Zambia               & 44.82                            & 86.20                            & 93.96                             & 2074                               & 231           \\
Spain                & 46.49                            & 77.19                            & 83.33                             & 2059                               & 227           \\ 
\bottomrule
\end{tabular}
\label{table:countrywise}
\end{table}

\begin{table*}[t]
\caption{Computational complexity and inference efficiency comparison of GeoQueryNet with baseline cross-view geolocalization methods on CVW500k.}
\begin{tabular}{lrrrrrr}
\toprule
\textbf{Models}      & \textbf{\begin{tabular}[c]{@{}r@{}}Total \\ Params (M)\end{tabular}} & \textbf{\begin{tabular}[c]{@{}r@{}}Trainable \\ Params (M)\end{tabular}} & \textbf{\begin{tabular}[c]{@{}r@{}}GFLOPs /\\ Batch(50)\end{tabular}} & \textbf{\begin{tabular}[c]{@{}r@{}}GFLOPs / \\ Image Pair\end{tabular}} & \textbf{\begin{tabular}[c]{@{}r@{}}Inference Time /\\ Batch (50) (ms)\end{tabular}} & \textbf{\begin{tabular}[c]{@{}r@{}}Inference Time /\\ Image Pair (ms)\end{tabular}} \\ 
\midrule
Sample4Geo\cite{deuser2023sample4geo} & 11.1765  & 11.1765  & 391.0309  & 4.0732   & 23.4351    & 0.4928                                                                              \\
GeoDTR\cite{zhang2023geodtr}          & 48.5086  & 48.5086  & 965.4937  & 48.2747  & 593.41   & 11.8682                                                                             \\
GeoDTR+\cite{zhang2024geodtrp}        & 30.0477  & 30.0477  & 209.7180  & 10.4859  & 167.405    & 3.3481                                                                              \\
CAMP\cite{wu2024camp}                 & 91.3954  & 91.3954  & 5784.2772 & 45.1896  & 471.035  & 9.4207                                                                              \\
DSTG\cite{liang2025dstg}              & 42.4401  & 42.4401  & 391.4572  & 12.2330  & 249.995  & 4.9999                                                                              \\
\textbf{GeoQueryNet} & 2188.0769 & 216.2878 & 4219.0629 & 527.3829 & 1102.17   & 22.0434                                                                             \\
\bottomrule
\end{tabular}
\label{tab:gflops_inference}
\end{table*}

\subsection{Analysis and Discussion}

\begin{figure}
    \centering
    \includegraphics[width=1\linewidth]{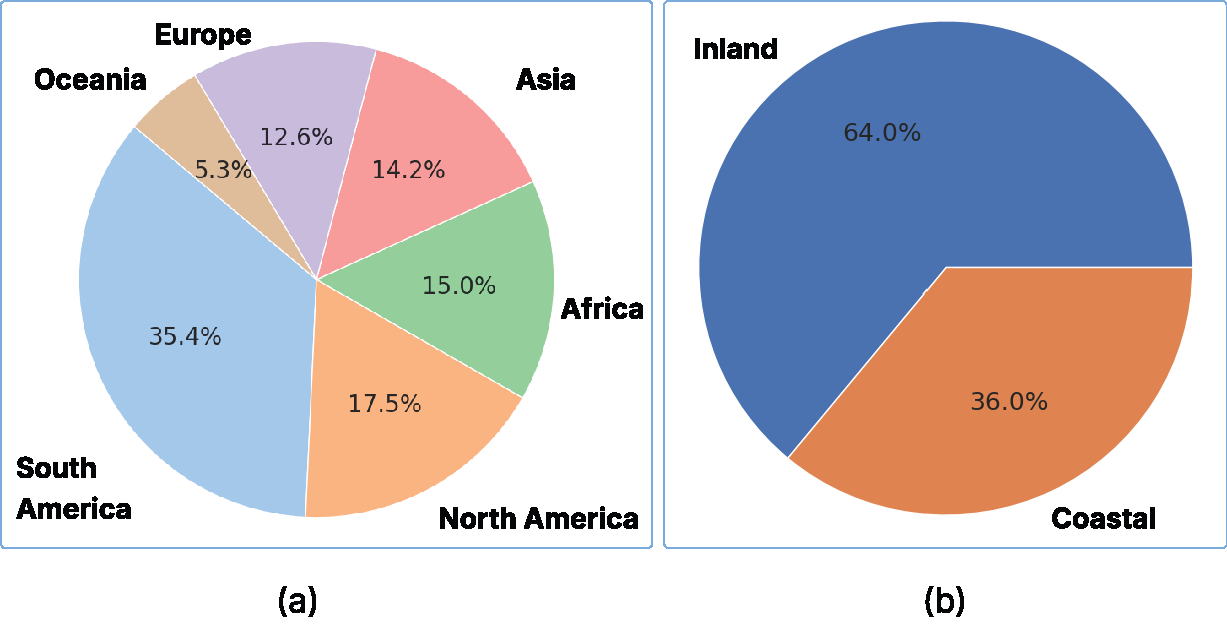}
    \caption{\textbf{(a) Continent distribution:} The pie chart illustrates the proportion of images collected from different continents. \textbf{(b) Coastal vs. inland distribution:} The dataset contains images from both inland and coastal regions.}
    \label{fig:pi_line1}
\end{figure}

\begin{figure}
    \centering
    \includegraphics[width=0.95\linewidth]{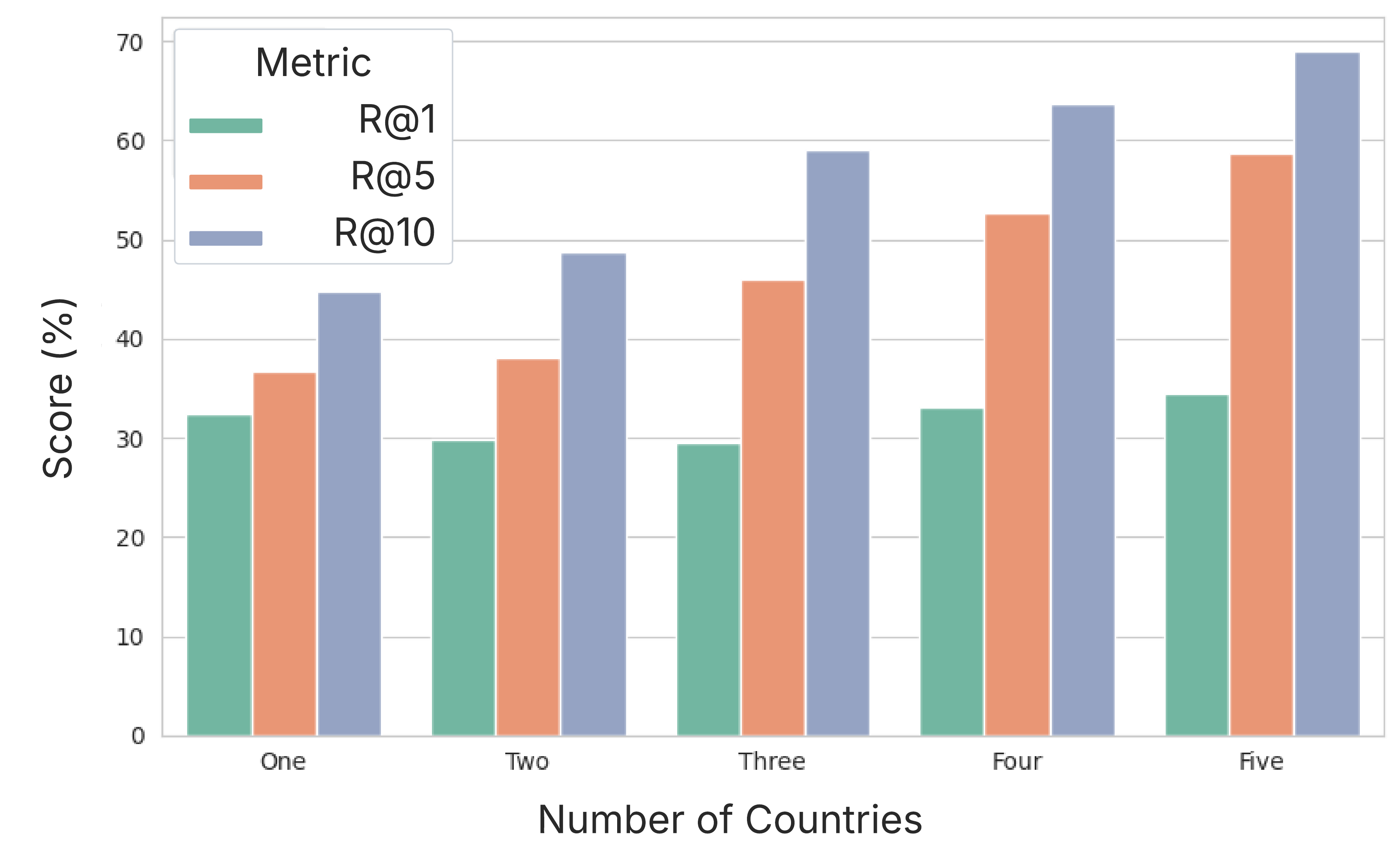}
    \caption{Impact of Geographic Diversity on Cross-View Retrieval Performance.}
    \label{fig:eval_num_country}
\end{figure}

To analyze the geographic diversity of the dataset, we examine both continental coverage and proximity to coastlines. As shown in Figure \ref{fig:pi_line1}(a), the dataset spans multiple continents, with the largest portion originating from South America, followed by North America, Africa, Asia, and Europe, while Oceania contributes a smaller but meaningful portion. 
\noindent
In addition, Figure \ref{fig:pi_line1}(b) presents a coastal versus inland analysis based on the distance of image locations from the nearest coastline. The results show that 64\% of the samples are located inland, while 36\% are from coastal regions. 



\noindent\textbf{Countrywise Geolocalization.}
Table \ref{table:countrywise} illustrates the geolocalization retrieval accuracy across ten countries using three evaluation metrics: Top-1, Top-5, and Top-10. Overall, Top-1 accuracy varies significantly between countries, ranging approximately from 26\% to 49\%, indicating the difficulty of correctly identifying the exact location from a single prediction. However, when the retrieval space expands to Top-5, the accuracy improves substantially for all countries, reaching values between 56\% and 86\%. Further expansion to Top-10 retrieval results in even higher performance, with several countries exceeding 80-90\% accuracy. Among the evaluated regions, Australia and Zambia consistently achieve the highest retrieval accuracy, suggesting that the model generalizes particularly well for these geographic environments. Countries such as Argentina and Russia exhibit comparatively lower Top-1 performance, although their accuracy improves significantly under Top-5 and Top-10 metrics. We show qualitative results and fail cases in the supplementary. 

\noindent\textbf{Impact of Geographic Diversity on Retrieval Performance:} Figure \ref{fig:eval_num_country} analyzes the generalization capability of the proposed model when evaluated on subsets of the CVW500k dataset containing different numbers of countries. The model is evaluated on five progressively reduced subsets where one country is removed at each step while keeping the number of data points constant. The results show a clear trend: retrieval performance consistently improves as the number of countries included in the evaluation increases. When only one country is evaluated, the model achieves approximately 32\% R@1, 36\% R@5, and 45\% R@10. As more countries are added, the performance gradually increases. This improvement suggests that the model benefits from greater geographic diversity, where varied environmental structures, and architectural patterns provide richer contextual cues that help the model distinguish different locations more effectively.

\noindent\textbf{Computational Cost.} Table \ref{tab:gflops_inference} compares the computational complexity and inference efficiency of the evaluated cross-view geolocalization models. Among the baseline methods, Sample4Geo is the most lightweight and efficient, requiring only 11.18M trainable parameters, 4.07 GFLOPs per image pair, and 0.49 ms inference time per image pair. GeoDTR has a moderate parameter count but incurs a much higher inference cost than GeoDTR+, indicating that the enhanced design of GeoDTR+ improves computational efficiency while reducing trainable parameters. CAMP has the highest parameter count among the baselines and also requires substantial batch-level computation, mainly reflecting its heavier feature extraction and contrastive attribute mining design. DSTG provides a more balanced trade-off, with moderate parameters and inference latency compared with the heavier baselines. In contrast, GeoQueryNet has the largest model capacity, with 216.29M trainable parameters, 527.38 GFLOPs per image pair, and 22.04 ms inference time per image pair. 
This indicates that GeoQueryNet is computationally more demanding than existing baselines, reflecting a design choice that prioritizes enhanced representational capacity and retrieval performance over computational efficiency and lightweight deployment.


\noindent\textit{\textbf{Limitations.}} Despite strong performance in world-scale cross-view geolocalization, our approach has key limitations. First, the dataset remains globally sparse, leaving many regions with limited or no ground-view coverage. Second, the alignment mechanism is sensitive to repetitive visual patterns in urban and rural scenes, such as similar roads, buildings, or agricultural layouts. As a result, the model may focus on visually plausible but incorrect satellite regions, causing ambiguous predictions, especially at world scale where distant locations can share similar structures.
\section{Conclusion}
\label{sec:conclusion}
In this work, we address the limitations of existing cross-view geolocalization research by introducing both a globally diverse dataset and a novel cross-view learning framework. We proposed \textbf{Cross-View World 500k (CVW500k)}, a large-scale dataset consisting of over 500,000 ground-satellite image pairs collected worldwide. Unlike previous benchmarks that focus on a limited number of cities or rely on panoramic ground views, \textbf{CVW500k} provides global geographic coverage and uses limited field-of-view ground images, a better representative of real-world localization scenarios.

To effectively handle the viewpoint gap and partial scene overlap between ground and satellite images, we introduced \textbf{GeoQueryNet}, a query-based cross-view fusion transformer architecture. 
This design enables the model to focus on spatial patterns that remain consistent across viewpoints.
Experimental results demonstrate that \textbf{GeoQueryNet} achieves strong performance on the proposed \textbf{CVW500k} dataset and significantly outperforms several existing cross-view geolocalization methods. Additional ablation studies further confirm the importance of both LoRA-based encoder adaptation and the Cross-View Alignment Module in achieving optimal performance.
Overall, the proposed dataset and framework provide an important step toward globally scalable cross-view geolocalization under realistic conditions. Future work may explore integrating temporal information, multi-modal cues, or large-scale pretraining strategies to further improve robustness and localization accuracy in complex real-world environments.


\bibliographystyle{ACM-Reference-Format}
\bibliography{sample-sigconf0}

\clearpage
\appendix
\setcounter{page}{1}

\section{Appendix}
\label{sec:Description}

In this supplementary, we have included additional analysis and comparison on the proposed dataset. We also show additional quantitative and qualitative results using the proposed approach. These results are important to support the manuscript and could not be included in the main paper due to the space constraint.\\
\textbf{Why we need a Limited Field-of-View Dataset?}
Widely used cross-view geolocalization datasets such as CVUSA, CVACT, and VIGOR have significantly contributed to the development of cross-view matching methods. Most of these datasets rely on panoramic ground images with an 360 degree field of view. Panoramic imagery often introduces geometric distortions due to the stitching and projection processes, a phenomenon that is particularly noticeable in datasets such as CVACT and VIGOR. These distortions can alter structural cues and lead models to learn dataset-specific artifacts rather than robust geographic features. In real-world applications, however, many images are captured with standard cameras that have a limited field of view, where only a small portion of the surrounding environment is visible. Consequently, models trained on panoramic datasets may struggle to generalize to such scenarios. A limited field-of-view dataset therefore provides a more realistic and challenging benchmark, requiring models to rely on distinctive local patterns and spatial structures rather than global panoramic context.





\begin{figure}[h]
    \centering
    \includegraphics[width=1\linewidth]{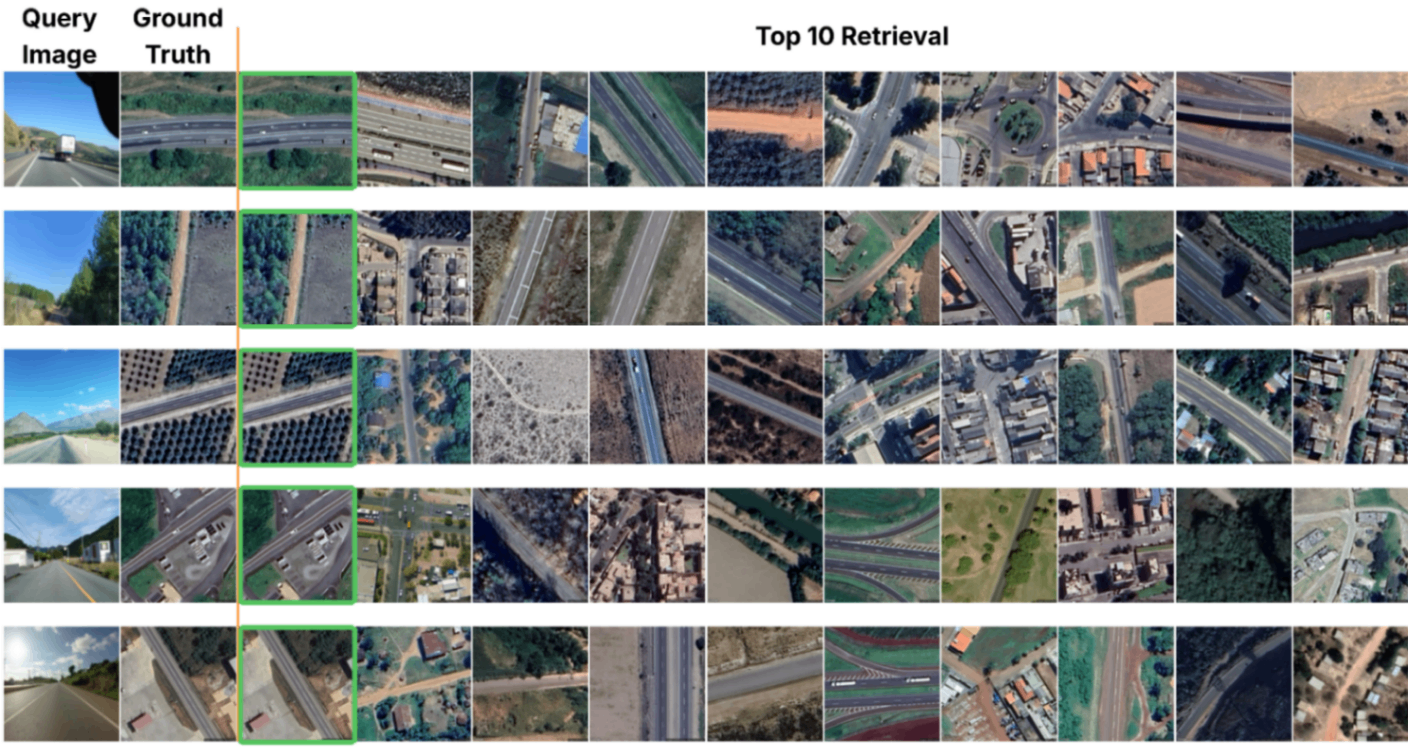}
    \caption{Success Case: Top 10 retrieval of our method on CVW500K dataset}
    \label{fig:succ_case}
\end{figure}

\begin{figure}
    \centering
    \includegraphics[width=1\linewidth]{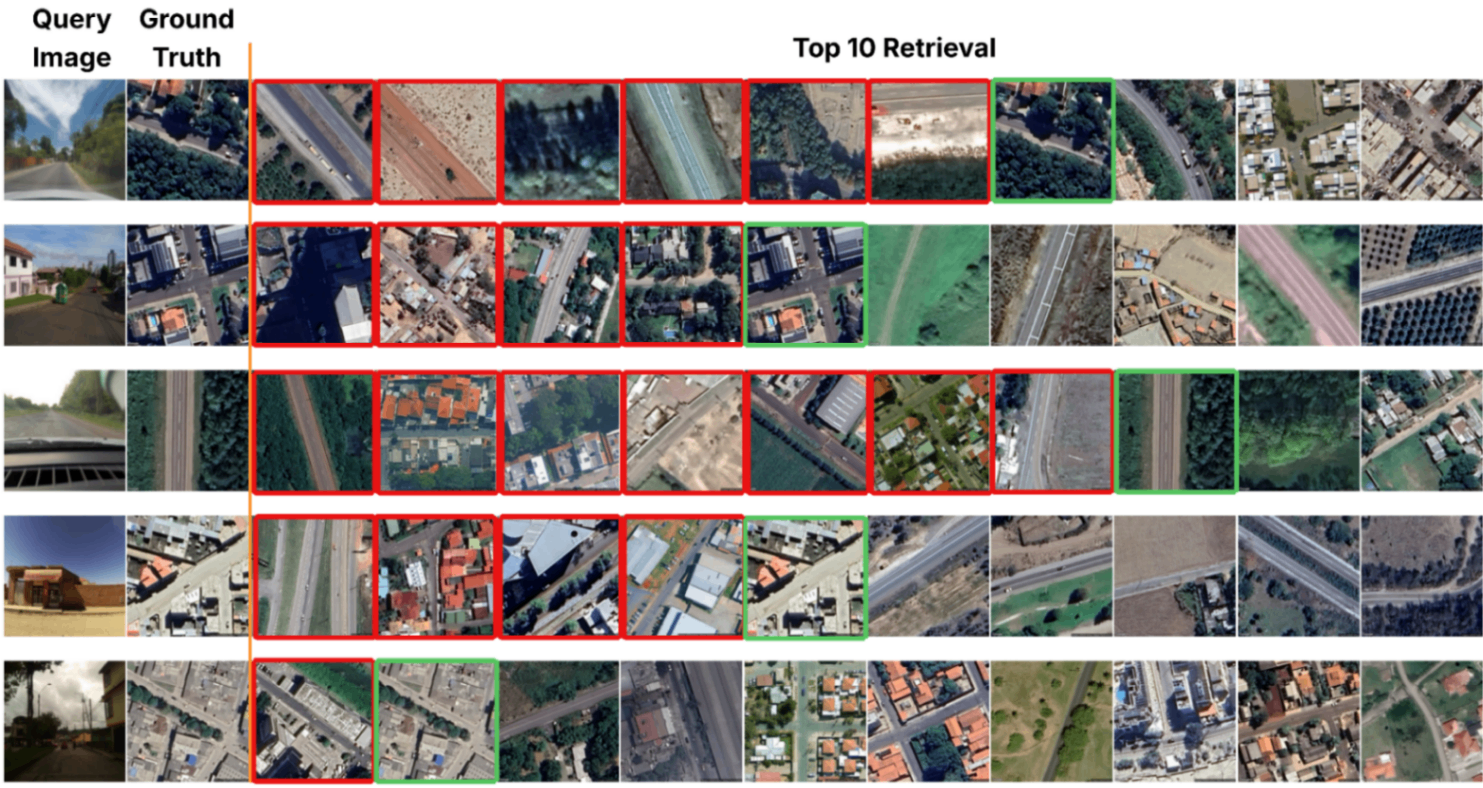}
    \caption{Partial Success Case: Top 10 retrieval of our method on CVW500K dataset}
    \label{fig:partial_case}
\end{figure}

\begin{figure}
    \centering
    \includegraphics[width=1\linewidth]{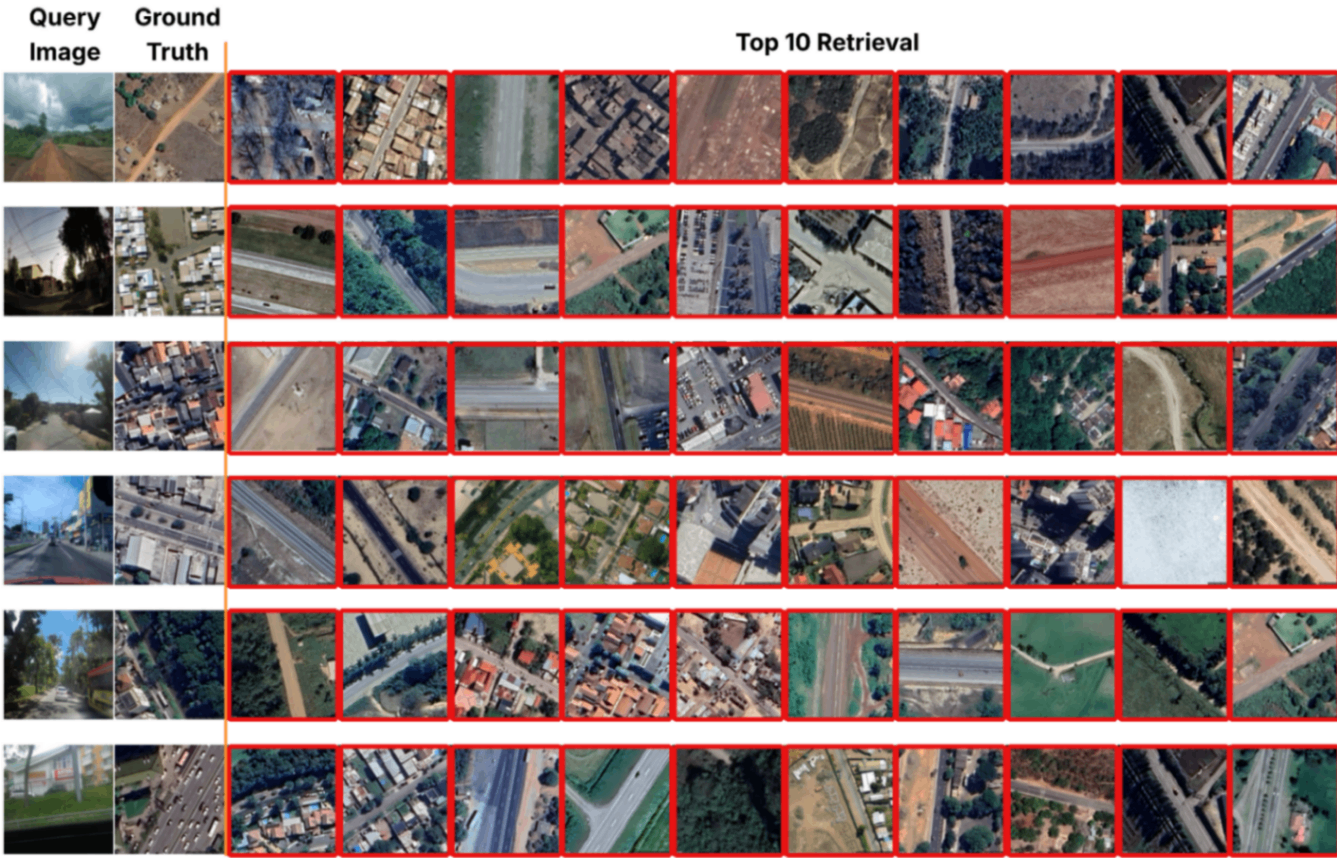}
    \caption{Failed Case: Top 10 retrieval of our method on CVW500K dataset}
    \label{fig:fail_case}
\end{figure}

\begin{figure}
    \centering
    \includegraphics[width=0.6\linewidth]{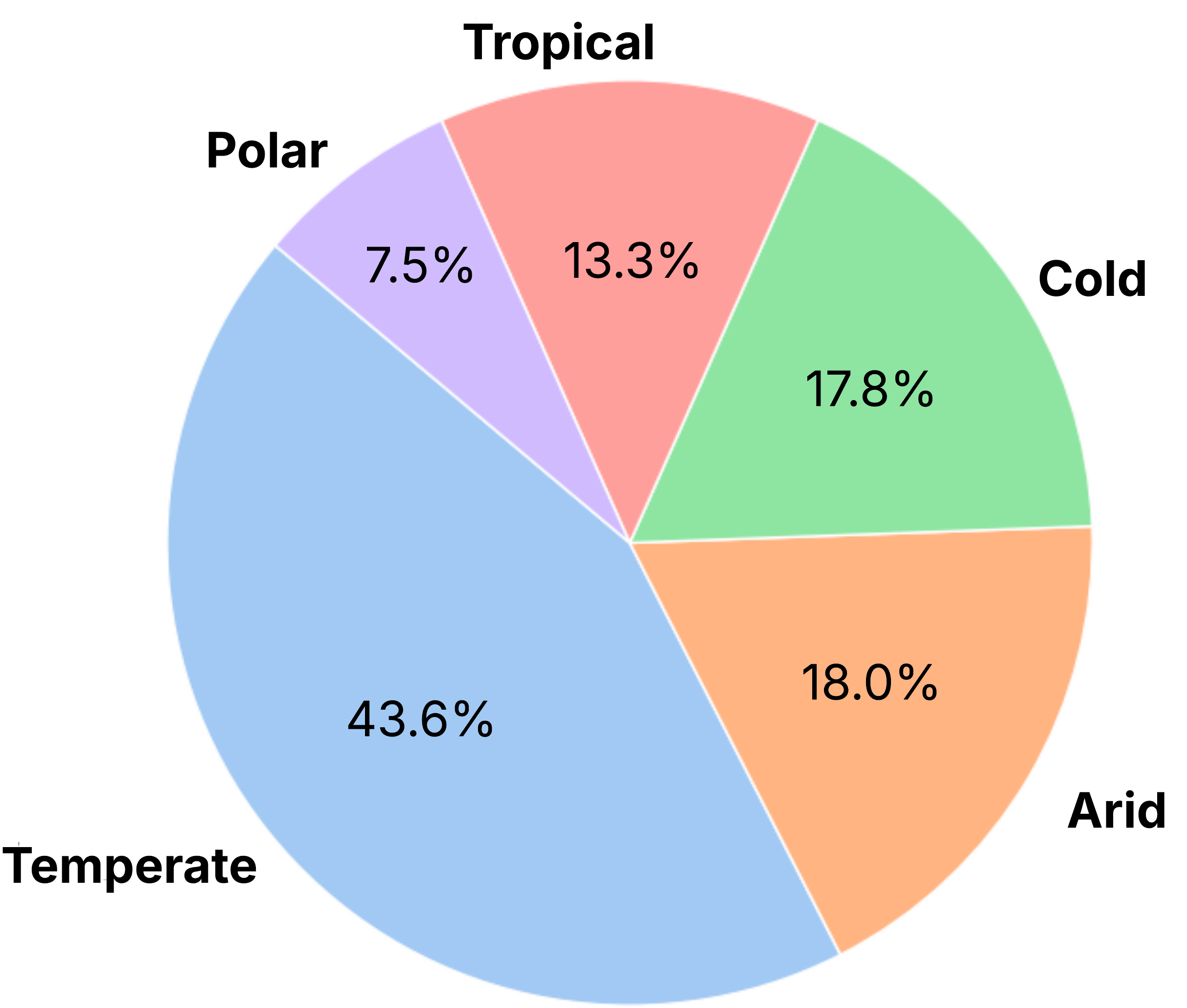}
    \caption{Climate zone distribution of the dataset based on the \textbf{Köppen climate classification}. The dataset covers diverse environmental regions including tropical, arid, temperate, cold, and polar climates, demonstrating broad geographic and environmental diversity.}
    \label{fig:climate_zone}
\end{figure}

\textbf{Climate Zone Coverage:}
In figure \ref{fig:climate_zone}, we categorize each location based on the Köppen climate classification\cite{Beck2018} to analyze the environmental diversity of the dataset. The results show that the dataset covers a wide range of climate zones, including tropical, arid, temperate, cold, and polar regions. This distribution indicates that the dataset captures diverse environmental conditions such as deserts, tropical forests, temperate urban areas, and cold northern landscapes, which is important for evaluating the robustness and generalization ability of geolocalization models.

\textbf{Successful and Failure Cases:}
Figures \ref{fig:succ_case}-\ref{fig:fail_case} illustrate qualitative results of our model evaluated on the CVW500k dataset, highlighting different levels of retrieval performance. Figure \ref{fig:succ_case} presents successful cases where the model correctly retrieves the corresponding satellite image for a given ground-view query. Figure \ref{fig:partial_case} shows partial success cases in which the retrieved results share similar geographic or visual characteristics with the ground image but do not correspond to the exact location. Figure \ref{fig:fail_case} illustrates failure cases where the model retrieves incorrect locations, often due to ambiguous visual patterns, or limited distinctive features in the scene.

\begin{figure}[H]
    \centering
    \includegraphics[width=0.99\linewidth]{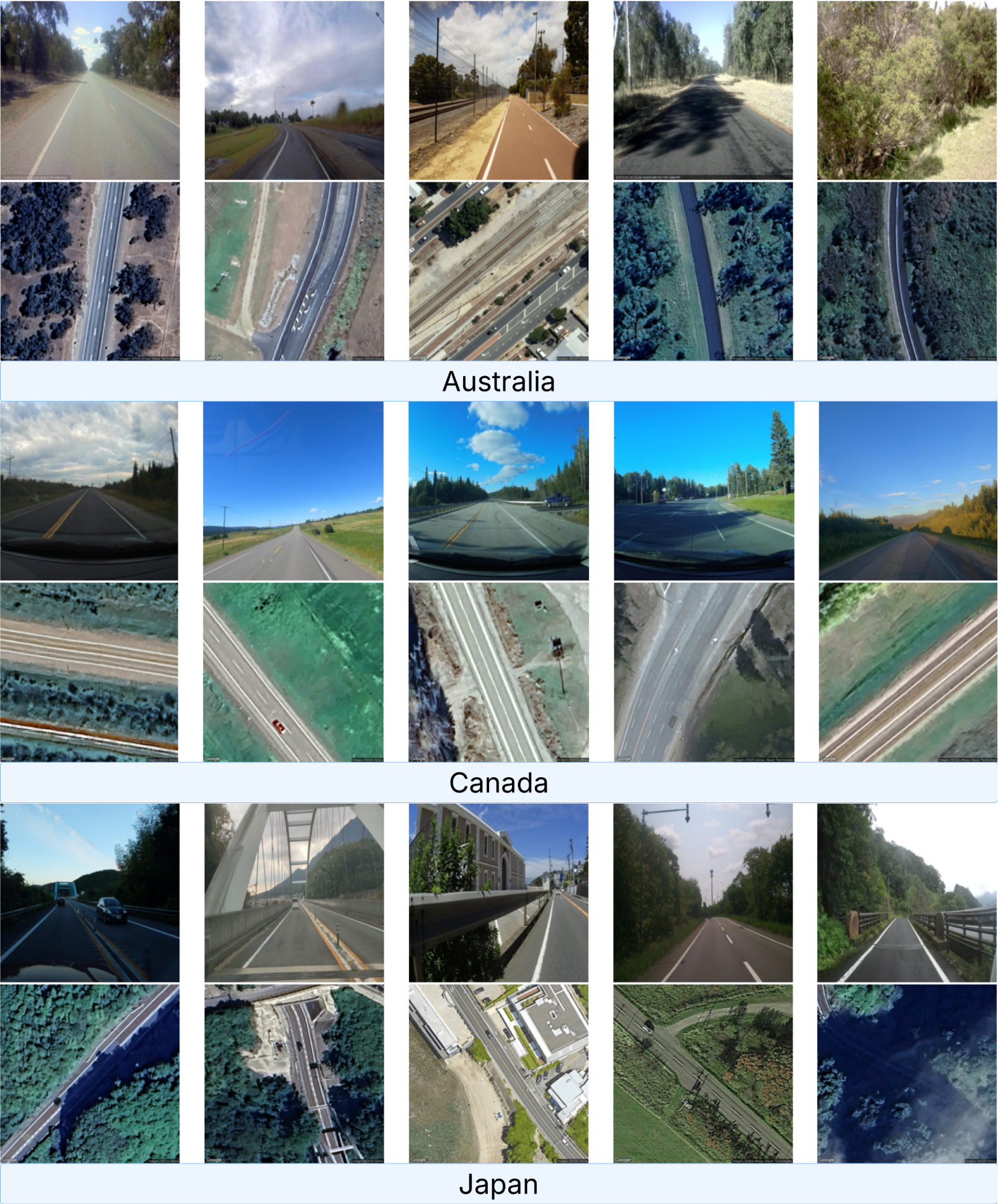}
    \label{fig:sample1}
\end{figure}

\begin{figure}
    \centering
    \includegraphics[width=0.99\linewidth]{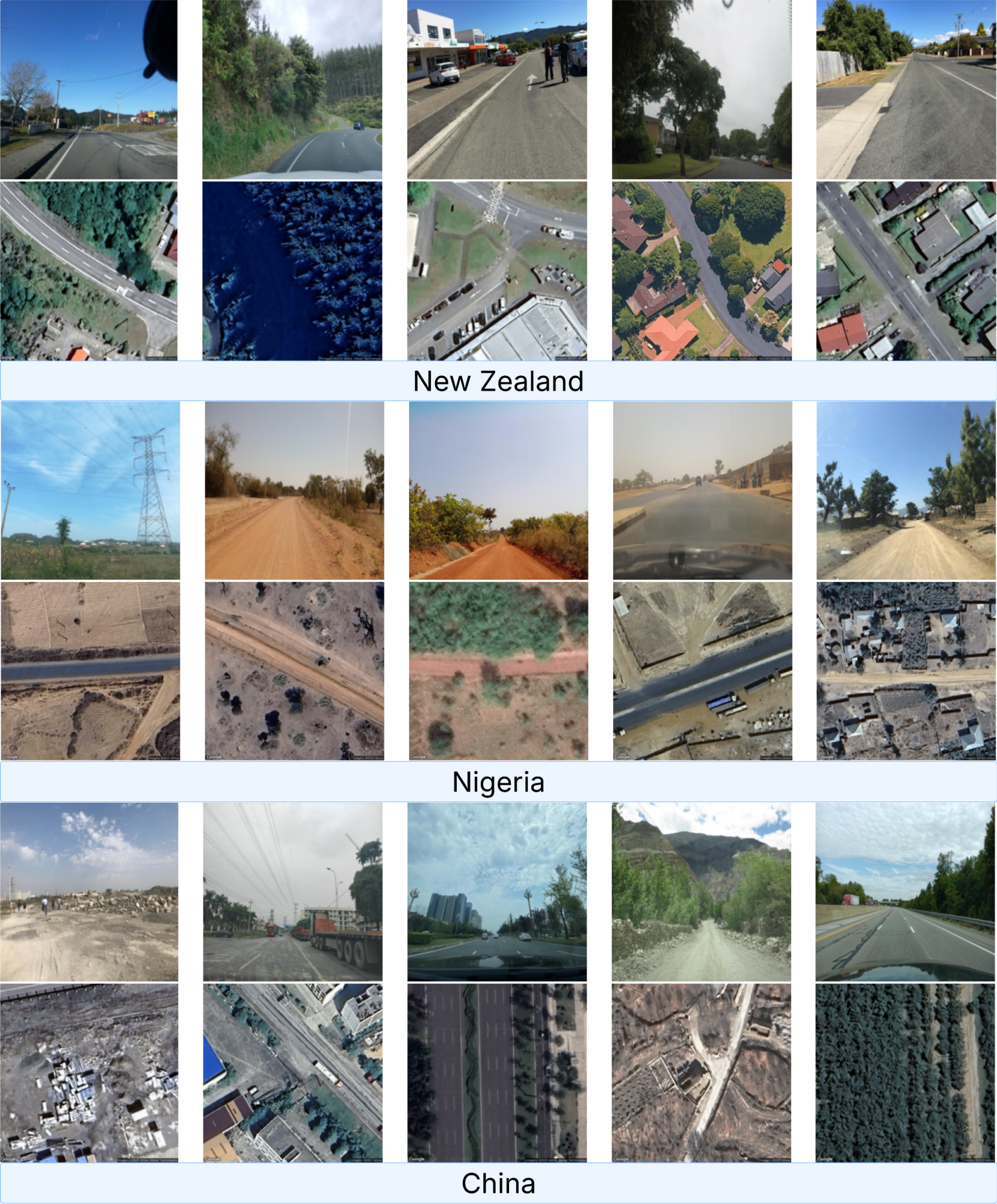}
    \label{fig:sample2}
\end{figure}

\begin{figure}
    \centering
    \includegraphics[width=0.99\linewidth]{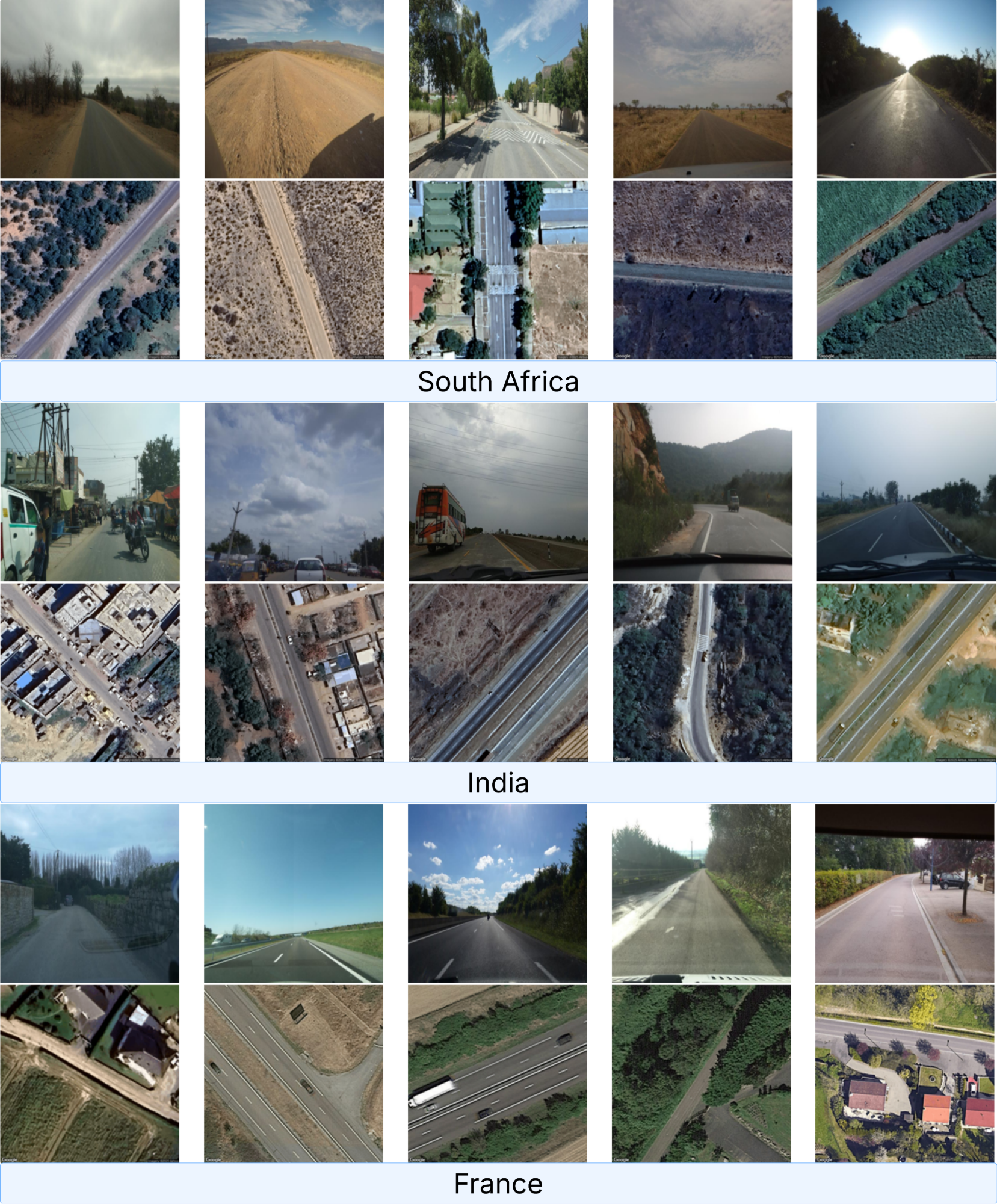}
    \label{fig:sample3}
\end{figure}

\vspace*{\fill}
\begin{center}
    \textbf{CVW500k} sample images from different countries.    
\end{center}


\end{document}